\documentclass[conference]{IEEEtran}
\usepackage{00_preamble}
\usepackage{00_macros}
\usepackage{00_spacing}

\begin{document}

\title{\LARGE \bf A Multi-Agent Security Testbed for the Analysis of Attacks and Defenses in Collaborative Sensor Fusion}

\newif\ifAnonymize

\Anonymizefalse

\ifAnonymize

\else
    \author{\IEEEauthorblockN{R. Spencer Hallyburton}
    \IEEEauthorblockA{Duke University\\
    spencer.hallyburton@duke.edu}
    \and
    \IEEEauthorblockN{David Hunt}
    \IEEEauthorblockA{Duke University\\
    david.hunt@duke.edu}
    \and\IEEEauthorblockN{Shaocheng Luo}
    \IEEEauthorblockA{Duke University\\
    shaocheng.luo@duke.edu}
    \and
    \IEEEauthorblockN{Miroslav Pajic}
    \IEEEauthorblockA{Duke University\\
    miroslav.pajic@duke.edu}}
\fi

\IEEEoverridecommandlockouts
\makeatletter\def\@IEEEpubidpullup{6.5\baselineskip}\makeatother
\IEEEpubid{\parbox{\columnwidth}{
    Network and Distributed System Security (NDSS) Symposium 2024\\
    26 February - 1 March 2024, San Diego, CA, USA\\
    ISBN 1-891562-93-2\\
    https://dx.doi.org/10.14722/ndss.2024.23xxx\\
    www.ndss-symposium.org
}
\hspace{\columnsep}\makebox[\columnwidth]{}}

\maketitle


\begin{abstract}
    The performance and safety of autonomous vehicles (AVs) deteriorates under adverse environments and adversarial actors. The investment in multi-sensor, multi-agent (\msma) AVs is meant to promote improved efficiency of travel and mitigate safety risks. Unfortunately, minimal investment has been made to develop security-aware \msma\ sensor fusion pipelines leaving them vulnerable to adversaries. To advance security analysis of AVs, we develop the Multi-Agent Security Testbed, \mast, in the Robot Operating System (ROS2). Our framework is scalable for general AV scenarios and is integrated with recent multi-agent datasets. We construct the first bridge between \avstack\ and ROS and develop automated AV pipeline builds to enable rapid AV prototyping. We tackle the challenge of deploying variable numbers of agent/adversary nodes at launch-time with dynamic topic remapping. Using this testbed, we motivate the need for security-aware AV architectures by exposing the vulnerability of centralized multi-agent fusion pipelines to (un)coordinated adversary models in case studies and Monte Carlo analysis.
\end{abstract}
\section{Introduction} \label{sec:1-intro}

Adversarial actors, sensor degradation due to adverse environments, and unavoidable natural occlusion threaten the prospect of safely deploying autonomous vehicles (AVs) in the real world. AVs that are ill-equipped for these challenging conditions are at risk for devastating safety-critical failures. The severity of such outcomes is apparent with recent real-world case studies: white-hat adversaries exposed that hackers could gain remote access to a vehicle's control center and manipulate messages on the CAN bus~\cite{miller2015remote}; a ``state-of-the-art'' commercially-available AV endured a high-speed collision due simply to sun reflecting off of a large vehicle and obscuring perception~\cite{boudette2021tesla}; natural occlusions were shown to cause delayed brake response time due to deteriorated motion prediction accuracy~\cite{lee2018collision}. Moreover, a recent report from the National Highway Traffic Safety Administration (NHTSA) described the increasing concerns in adversarial remote exploitation of vehicle hardware and software and illuminated the lack of security protocols that could mitigate threats~\cite{nhtsa2016}. 

Challenging environments and adversarial actors are formidable barriers for a single agent to overcome. An important avenue for mitigating adversity is to pursue multi-sensor, multi-agent (\msma) collaborative autonomy. Multi-sensor systems leverage multiple sensing modalities including but not limited to cameras, \lidars, and \radars. Multi-agent systems share information with external platforms via vehicle-to-vehicle (V2V) and vehicle-to-infrastructure (V2I) channels. Such \msma\ concepts are essential to the security awareness of perception, planning, and control because they provide complementary perspectives and add necessary redundancy over single-source systems.

Government and industry have been investing heavily in \msma\ vehicle technology for over a decade. Experts have used this investment to foreshadow high operating ranges~\cite{5gaa5GV2XOnward}, low latency~\cite{5GV2X_vs_IEEE_802_11p}, and robust physical testing infrastructure for academic research~\cite{MCity_about, MCity_vehicle_sensors}. The investment in connected vehicle technologies is meant to improve travel efficiency and mitigate immediate safety risks. Unfortunately, it coincides with a dramatic rise in cyber threats against cyber-physical systems (CPS)~\cite{zou2021cyber}. In particular, the confluence of demonstrated remote attacks on AVs with the desire to expand networked connectivity of AVs suggest an impending increase in cyber threats against~AVs that could compromise \msma\ pipelines.

Despite the investment in connected vehicles, \textit{minimal effort has been made to develop security-aware \msma\ collaborative algorithms}. To spark development in this critical area, we design and implement a scalable \textbf{multi-agent security testbed, \mast}, based on v2 of the Robot Operating System (ROS2~\cite{quigley2009ros}). With \mast, an evaluator can spin up multiple mobile and static (infrastructure) agents and evaluate \msma\ pipelines under benign and adverse conditions including security threats. \mast\ is released open-source\footnote{
\textit{redacted}
}.

\mast\ uses open-source datasets to feed \msma\ sensor data to agent nodes in an event-driven simulation. The recently-released \msma\ dataset generation pipeline from~\cite{hallyburton2023datasets} provides a general framework to generate limitless \msma\ data from complex scenes with multiple mobile and static agents. We integrate the datasets from~\cite{hallyburton2023datasets} and illustrate their effectiveness in exploring multi-agent security case studies.

\mast\ also uses the advanced control algorithms and perception models developed for AVs in a real-time manner, which are provided by \avstack~\cite{avstack}, an emerging platform for the design, implementation, testing, and analysis of AVs. However, prior to this work, \avstack\ was only used in an offline, postprocessing context. To integrate \avstack\ with the near-real-time environment of \mast, we leverage the real-time capabilities of ROS and develop the first bridge between \avstack\ and ROS. This bridge is released open-source\footnote{
\textit{redacted}
}.

In \mast, we bridge the gap between AV pipelines and the module menagerie implemented in \avstack\ by developing automated pipeline builds. The build process uses \textit{registries} inspired by OpenMMLab~\cite{2020mmdet3d}. Prior to execution, the designer will specify the configuration of the AV pipeline in a high-level language in \mast. At launch-time, the build process will build each module of the AV pipeline by searching for algorithms in the registries. This process simplifies executing case studies and Monte Carlo analysis and yields faster deployment. 

To advance security analysis of \msma\ autonomy in \mast, we implement baseline adversary nodes capable of disrupting sensing, perception, or tracking operations in the AV pipeline. We consider both uncoordinated and coordinated models of multi-adversary attacks on multi-agent systems. In the former, decisions of each adversary are independent. In the latter, adversaries can collude to consistently compromise multiple agents. To consider realistic scenarios, we consider that only a subset of agent nodes are compromised by adversaries. To handle a variable number of agents and adversaries at launch-time, we create a dynamic topic remapping approach that allows us to insert a number of adversaries into the agent pipelines without modifying downstream algorithms/nodes.

As an evaluation, we provide case studies with multiple adversary models on multi-agent datasets to illustrate the effectiveness of \mast\ in performing security analysis. We explore how (un)coordinated adversaries can affect multi-agent sensor fusion through in-depth case studies with visualizations of sensory, perceptive, and tracking data for each agent. Videos of case studies are released online\footnote{\url{https://sites.google.com/view/cpsl-mast/home}}. Finally, we demonstrate running a Monte Carlo analysis by scripting over \mast\ to capture high-level insights into the robustness of existing fusion pipelines to adversarial actors. We find that the lack of security awareness in vanilla sensor fusion pipelines leaves collaborative autonomy dangerously vulnerable to exploitation.

In summary, this work makes the following contributions:
\begin{itemize}
    \item We design \mast, the multi-agent security testbed, using ROS2 for security-analysis of \msma\ autonomy.
    \item We integrate \mast\ with the recently-released multi-agent dataset pipeline~\cite{hallyburton2023datasets} for the first real-time playback of simulated \msma\ autonomy data.
    \item We build the first bridge between \avstack~\cite{avstack} and ROS to integrate state-of-the-art AV algorithms into a scalable, near-real-time simulation.
    \item We design an automated build process for AV algorithm pipelines using high-level configuration files to enable rapid prototyping of \msma\ configurations.
    \item We provide a baseline framework for security analysis of \msma\ algorithms in \mast\ with (un)coordinated models of adversary nodes that can affect the sensing, perception, tracking, and/or communication of AVs.
    \item We execute security-focused case studies and Monte Carlo evaluations with centralized sensor fusion and motivate the need for security-aware collaborative autonomy.
\end{itemize}



\section{Multi-Agent Collaboration} \label{sec:2-multi-agent-avs}

Due to the \textit{safety-critical} nature of AVs and the vulnerabilities induced by opening wireless communication channels, safety-critical features such as adaptive cruise control, lane centering, and obstacle avoidance will be influenced primarily by local information. Multi-agent collaboration will play a significant role in the validation of local information and in \textit{mission-critical} functions such as broad situational awareness, path-planning, and urban maneuvering.

\begin{figure}[t]
    \centering
    \begin{subfigure}[b]{0.8\linewidth}
        \includegraphics[width=\linewidth]{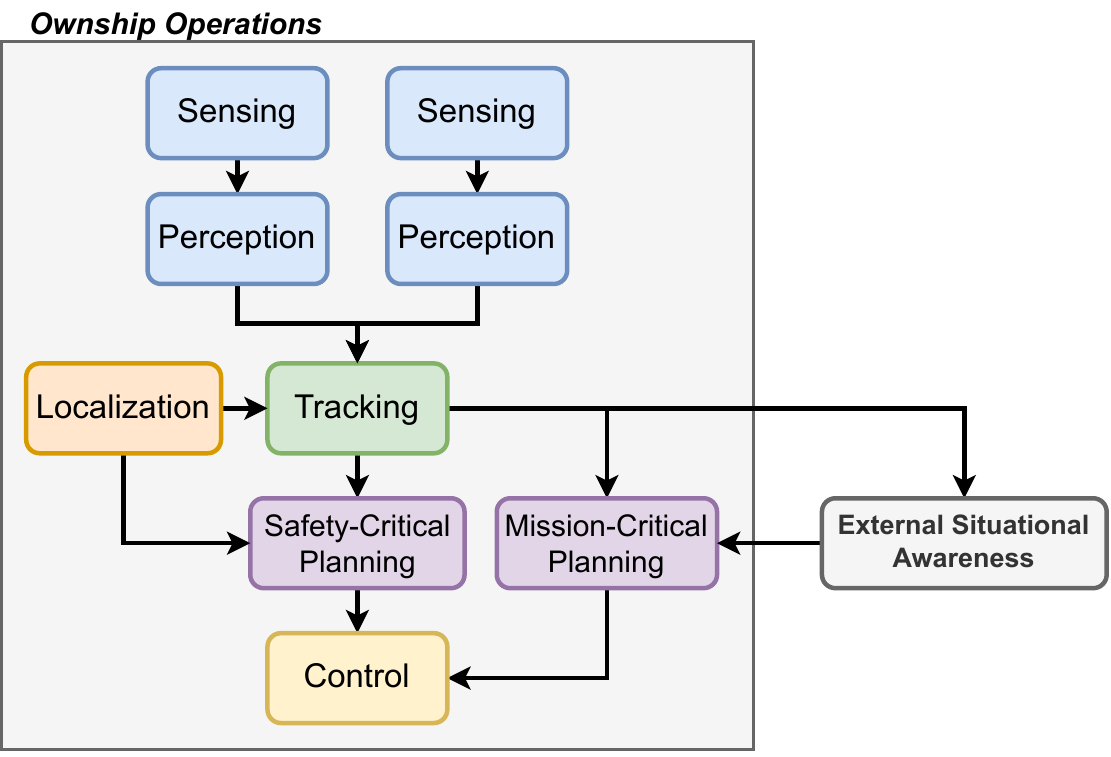}
        \caption{Ownship operations for each agent fusing external awareness for mission-critical planning.}
    \end{subfigure}
    \begin{subfigure}[b]{0.8\linewidth}
        \includegraphics[width=0.9\linewidth]{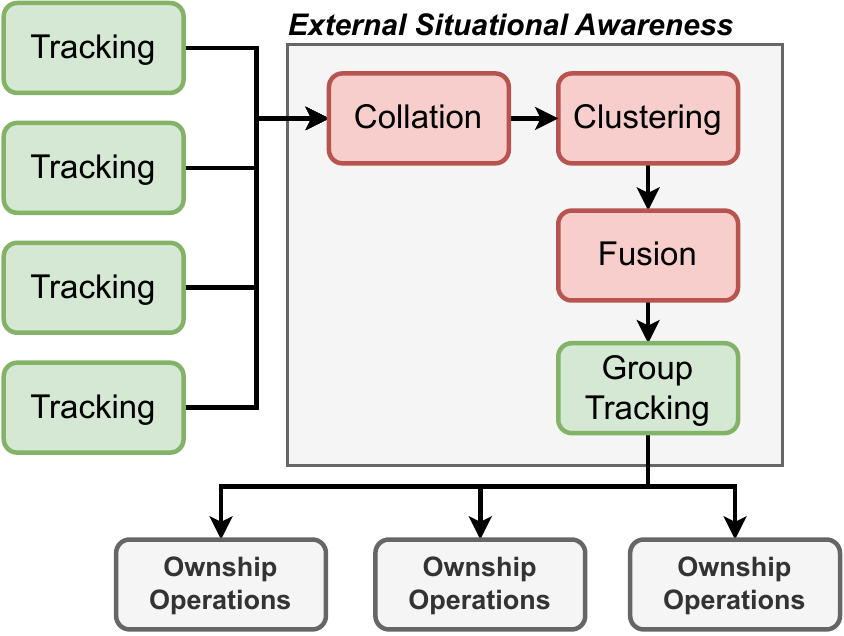}
        \caption{Centralized fusion via command center to provide unified operating picture across multiple agents.}
    \end{subfigure}
    \caption{(a) Ownship operations use local information for safety-critical planning and incorporate external knowledge for mission-critical. Safety-critical is partitioned to mitigate impact of adversarial actors in the sensing network. (b) Command center integrates situational awareness from all agents. Collation synchronizes data while clustering, fusion, and group tracking distill agents' tracks into a unified operating picture.}
    \label{fig:multi-agent-pipeline}
\end{figure}

Both on-board and collaborative computation will occur. We choose to partition each agent's sensor fusion pipeline into two estimators: (1) safety-critical estimator that depends only on ownship information to ensure the safety and security of essential tasks, and (2) a mission-critical estimator that incorporates information from both local and collaborative/distributed systems. Fig.~\ref{fig:multi-agent-pipeline} demarcates the roles of onboard and distributed computation in a multi-agent system. Information is passed between the local and external systems via wired/wireless communication channels. We consider centralized collaboration where a ``command center'' (\commandcenter) operates as an edge server that fuses information from multiple agents.

\subsection{Safety-Critical Sensor Fusion}
AV operations such as localization, lane centering, adaptive cruise control, and obstacle avoidance are safety-critical functions for the AV. To mitigate the impact of an adversary at the communication channels or in the unknown network of ``collaborative'' systems, these functions may rely only on local information. In this model, each agent assumes he is not compromised and that local information is sufficient to maintain safety-critical levels of performance.

\subsection{Mission-Critical Sensor Fusion}
Many of the functions in an AV are mission-critical. Failure of a module will impact the quality of service in the form of traffic delays, erroneous/compromised data reconnaissance, improper routing, erratic behavior, or an inability to move due to a forced system shutdown. Modules including high-level route optimization and medium-scale path planning via an awareness of nearby object states are considered mission critical and can benefit from collaborative sensor fusion. 

\subsection{Implementation Details}
Here we present the implementation details of the agent and \commandcenter\ algorithms. For simplicity, we fix these pipelines throughout the case studies and Monte Carlo evaluations of Section~\ref{sec:6-evaluation}. As illustrated in Section~\ref{sec:av-pipelining}, rearchitecting agent and \commandcenter\ algorithm pipelines is a simple process with \texttt{config}-based builds and will be included in future works. We consider a scenario where agents are estimating states of objects (e.g.,~other vehicles) in the environment and the \commandcenter\ is fusing object states gathered from the agents.

\subsubsection{Dataset}
The case studies and results in this work use the CARLA infrastructure dataset~\cite{hallyburton2023datasets}. The dataset consists of one mobile agent (``ego'') and many static agents  (``infrastructure''). The ego is connected to the ground plane via the vehicle chassis. The infrastructure agents are placed at dedicated positions in the scene to emulate a smart-city configuration. The infrastructure agents reside at 15~m elevations with a $30^{\circ}$ downward pitch for an advantageous vantage point.

\subsubsection{Agent Pipelines}
Each agent has its own pipeline to filter and process ownship sensing data. The agent then sends estimated object states to the \commandcenter\ to be collated and fused. This model of sending high-level tracked states is consistent with the bandwidth and latency requirements of existing and future V2X systems~\cite{5gaa5GV2XOnward}.

\paragraph{Sensing.} Each agent is equipped with a 10~Hz forward-facing RGB camera with a $90^{\circ}$ field of view, a 1.4 fstop, and a 5~ms shutter. These parameters can be modified during dataset generation~\cite{hallyburton2023datasets}. Each agent also possesses a rotating 64-channel \lidar\ sensor time-synchronized with its camera. The ego's \lidar\ has an elevation span of $[-17.6^{\circ},\,2.4^{\circ}]$ ($0^{\circ}$ is straight forward) with full azimuth coverage. The infrastructure \lidar\ spans $[-25^{\circ},\,30^{\circ}]$ with $90^{\circ}$ azimuth coverage to provide a hybrid forward/bird's-eye-view.

\paragraph{Perception.} Sensor data are stored in datasets meaning the agents paths cannot be altered during \mast\ evaluation. To take advantage of this ``replay'' context for efficiency, we apply perception to \lidar\ data offline and generate detections prior to running \mast. We use models provided by OpenMMLab~\cite{2020mmdet3d} via \avstack~\cite{avstack}. During the evaluations, we replay the detections in lieu of raw sensor data since the agent dynamics are fixed.

\paragraph{Tracking.} Detections are instantaneous estimates of object state. This state typically consists of positioning and sizing information. To build a longitudinal estimate of object state, detections are passed to multi-object tracking algorithms that score for object existence probabilities and associate incoming detections with existing tracked objects. Longitudinal states are necessary for prediction and planning. We use multi-object tracking algorithms provided by~\cite{avstack}. These are classical algorithms using bipartite matching (e.g.,~\cite{1986JVC}) for assignment and extended Kalman filtering for state estimation (e.g.,~\cite{1986blackmanRadar}).

\subsubsection{Command Center Pipeline} \label{sec:2-multi-agent-avs-cc}

The advent of V2X communication and the investment in smart-city networking technologies will enable computation at edge server nodes~\cite{MCity_C_V2X_Demo}. We model that mobile and static agents are connected via wireless/wired interfaces, respectively. This configuration is motivated by the University of Michigan's MCity~\cite{MCity_C_V2X_Demo}. In this configuration, infrastructure agents send estimated object states to the \commandcenter\ over wired connections while mobile agents experience higher latency and lower bandwidth over wireless (e.g., C-V2X or 802.11p transceivers~\cite{5gaa5GV2XOnward,MCity_C_V2X_Demo,Audi_Duesseldorf_Demo}). 

\paragraph{Collation.} In a complex network of agents, messages will arrive asynchronously. While \mast\ can replay data either synchronously or asynchronously, perfect synchronicity is infeasible in the real world. To handle the practical case of asynchronous transmission, the \commandcenter\ maintains a circular, min-heap priority queue for each agent in the network to collate the data. At a predefined and tunable latency factor, the collation algorithm will determine the earliest viable timestamp and pop a batch of track states closest to that time from each of the agents. Each of the tracks is transformed such that all tracks are in a common frame of reference (e.g.,``world'' frame). 

\paragraph{Clustering.} Agents with overlapping fields of view may each obtain a track on the same physical object. However, the agents will not intrinsically know these tracks follow the same object. To merge common track states across the multiple agents, the \commandcenter\ clusters tracks from different agents using the tracks' position states. The clustering algorithm is provided by \avstack~\cite{avstack} and is a simplified case of the classical DBSCAN algorithm. The output is a set of track clusters that each aspire to represent a single, unique object in the scene. 

\paragraph{Fusion.} Depending on the agents' field of view overlap, the clusters may have one or more members where each member is from a different agent. The state estimates from each of the members are combined in the \textit{fusion} step to distill a fused state estimate for each cluster. Since each member is a track and tracks have temporal correlations, we fuse member states using track-to-track fusion with the covariance intersection algorithm~\cite{2017ddfwithCI}. After fusion, each cluster has a representative state in addition to its track members.

\paragraph{Group Tracking.} Clustering/fusion does not consider the temporal consistency of clusters. Thus, clusters lack longitudinally-consistent labels and existence probability scores. We use group tracking at the \commandcenter\ to pass fused estimates of cluster states as detections to tracking algorithms. This approach is a special case of the widely-studied field of \textit{group tracking} that has extensive theory and algorithms~\cite{1986blackmanRadar}. The output of tracking is the \commandcenter's estimate of the state of unique objects in the scene. These tracks are then fed back to each of the agents as mission-critical situational awareness.
\section{Multi-Agent Security Testbed (MAST)}

\mast\ is built on ROS2 to leverage the strong community of open-source robotics. We design \mast\ to be scalable for pursuing many-agent, many-sensor evaluations with both distributed and centralized fusion architectures. Here, we describe the data sources, ROS nodes, and algorithm pipelines used to spin up \msma\ evaluations in \mast.

\subsection{Multi-Agent Data Sources}

Datasets from the recently-released multi-agent dataset generation framework~\cite{hallyburton2023datasets} drive event-based simulations in \mast. The multi-agent datasets contain multi-modal perception data and ground truth object state information from static and mobile agents. In an event-based simulation, the playback rate can scale to support the required computation from each of the nodes; \mast\ can run faster or slower than the wall-clock time depending on the computation needs. In this way, \mast\ is not limited to a maximum number of agents/sensors.

\subsection{ROS Nodes}
\noindent The ``node'' is the base execution process in ROS. Nodes communicate messages across topics. Here, we describe the relevant nodes used to build \mast\ environments.

\paragraph{Simulator} The simulator node drives the event loops. Agent pose and sensor data are read from the CARLA multi-agent dataset~\cite{hallyburton2023datasets} and published at the desired rate. Ground-truth object information is also published.

\paragraph{Mobile Agent} The CARLA dataset captures data from mobile agents spawned in the town. To simplify the experiments, we consider an infrastructure-only case where there is a single mobile agent (the ``ego''). 

\paragraph{Static Agents} Infrastructure sensing is an important pillar of the future of multi-platform collaborative sensing. We spawn nodes for multiple infrastructure sensors nearby to the ego vehicle for V2X collaboration.

\paragraph{Command Center} The \commandcenter\ is composed of broker and primary nodes. The broker polls to collect data from each of the agents and performs collation as described in Section~\ref{sec:2-multi-agent-avs-cc}. The broker sends collated data to the primary node; the primary node executes the remainder of the \commandcenter\ pipeline.

\paragraph{Adversary} We model the network as if each agent were susceptible to an adversary. Uncoordinated adversaries attach to the compromised agent nodes directly. Coordinated adversaries live in the communication network and communicate with a special coordinator node that is responsible for synchronizing attack directives among adversaries.

\begin{figure}
    \centering
    \begin{subfigure}[b]{0.54\linewidth}
        \fbox{\includegraphics[width=0.95\linewidth,trim={7cm 6cm 4cm 0},clip]{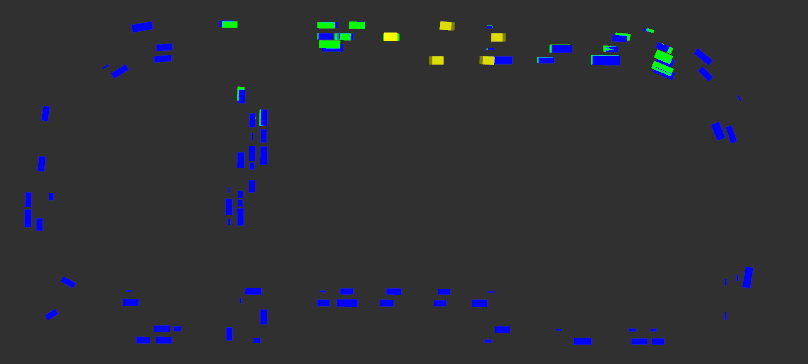}}
        \caption{ROS visualizer with objects.}
        \vspace{2ex}
        \fbox{\includegraphics[width=0.95\linewidth]{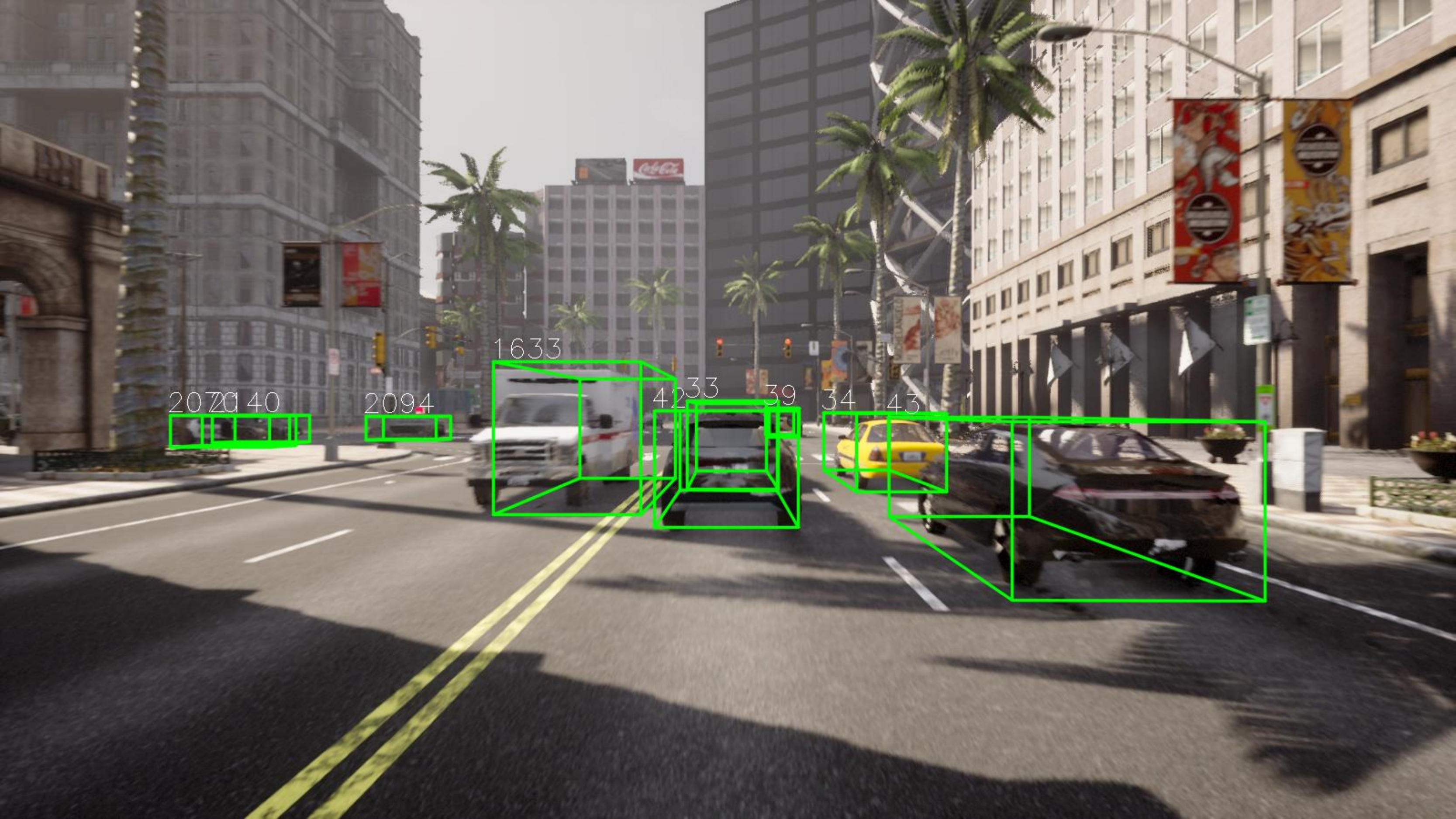}}
        \caption{Ego image with objects (green).}
    \end{subfigure}
    \begin{subfigure}[b]{0.42\linewidth}
        \fbox{\includegraphics[width=0.95\linewidth,trim={2cm 2cm 5cm 4cm},clip]{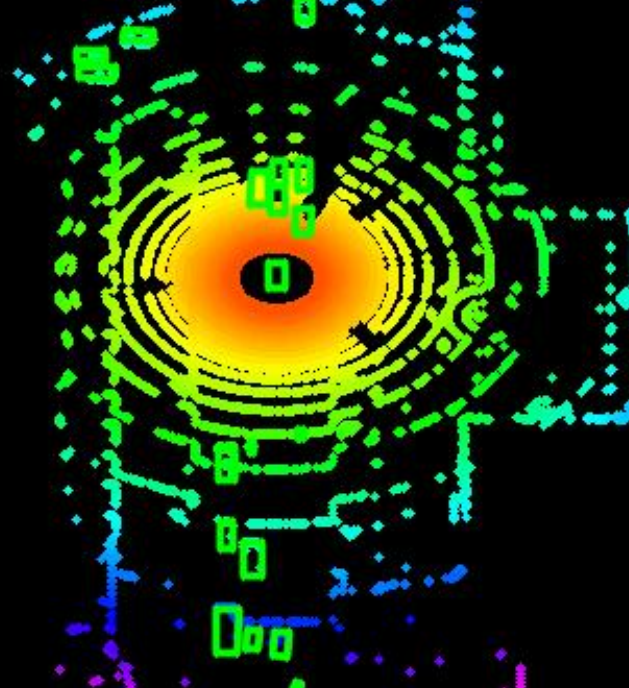}}
        \caption{LiDAR BEV with command center objects (green).}
    \end{subfigure}
    
    \caption{(a) Bird's-eye-view of multi-agent simulation under an unattacked scenario. Tracked object state estimates (green) overlap with ground truth object locations (blue). Agents (yellow) are positioned in the scene for case-study purposes and clearly do not provide full coverage over the town. Four infrastructure agents and one mobile agent (``ego'') provide object estimates to the \commandcenter. All tracks are pushed to the visualizer for real-time feedback. (b, c) Image and \lidar\ data, respectively, from ego agent with tracked objects from command center overlayed in green.}
    \label{fig:qt-visualizer}
\end{figure}

\paragraph{Visualizer} A QT-based visualizer runs to update the operator on the performance in real-time; an example frame is shown in Fig.~\ref{fig:qt-visualizer}. The visualizer subscribes to information furnished by the agents and is a useful debugging tool.

\subsection{Launch-Time Flexibility and Scalability}

Many multi-agent/security evaluations are tailored to a particular dataset or agent configuration. For example~\cite{zhang2023data} generates an adversarial dataset, Adv-OPV2V~\cite{2022openv2v}, with adversary optimization prior to execution. Similarly,~\cite{conte2018development} develops a ROS framework for evaluating surface vessels but restricts operation to a single ego agent and a fixed number of agents a-priori. In contrast, \mast\ is scalable to an arbitrary number of agents and adversaries. Moreover, \mast\ is flexible enough to handle distributed and centralized sensor fusion and to handle a diversity of agent and \commandcenter\ algorithm pipelines. The scalability and flexibility of \mast\ were achieved via the following innovations to the ROS launch configuration files.

\paragraph{Programmatic Node Launches}
A requirement for a flexible multi-agent evaluation framework is to specify the number of agent nodes at launch-time rather than hard-code the agents at build-time. We leverage recent advancements in ROS2 to design flexible launch files in a `bringup' package. The launch files use \texttt{OpaqueFunctions} to defer scenario configurations to launch-time rather than the standard process of setting at built-time. This allows users to dynamically set the number of agent and adversary nodes.

\paragraph{Data Structures and Discovery Functions}
Multi-agent networks will experience range limitation, quality of service (QoS) variability, and packet drops. Unfortunately, many research works set a fixed number of agents for each run and assure via software that each agent communicates with zero latency and no drops. Many of the algorithms developed in these overly clean environments are ill-suited to handle the variability experienced in the real-world, i.e., they break in response to natural variability. To improve robustness of multi-agent evaluations, all data containers are circular, min-heap priorities queues to handle out-of-order or lost data. Moreover, the \commandcenter\ broker runs a discovery function that receives agent status messages. This allows the network to expand and contract without breaking data structures and algorithms.

\paragraph{Dynamic Topic Remapping}
In a general multi-agent analysis or a specific security analysis, it can be desirable to apply a function to a \textit{subset} of the agent nodes (e.g.,~an adversary attacks sensor data of a single agent). To intercept inputs/outputs of an agent node and apply functions for any reason, we develop a dynamic ROS topic remapping solution for \mast. The launch file will dynamically remap the topic to pass through any specified intermediate functions before that topic is published for all subscribers. The downstream subscribers/nodes will not observe any accounting differences when topics are remapped through intermediaries. This feature was essential in developing a scalable and reconfigurable multi-adversary, multi-agent framework.

\subsection{Bridge to \avstack}

\avstack\ unifies and streamlines the AV design and prototype process~\cite{avstack}. Unfortunately, it was previously limited to non-real-time applications because of the Python global interperter lock (GIL). This meant that \avstack\ was only useful as a postprocessing and analysis tool. To support development of near-real-time AVs and to enable more flexible evaluations, we develop the first bridge between ROS and \avstack. \avstack\ provides libraries of tested algorithms from the open-source community for AV design while ROS defines a framework for scalable multiprocessing and communication; the bridge enables automatic conversion of datatypes and messages between the two. The bridge has been released open-source\footnote{
\textit{redacted}
} to support future AV development and evaluation.

\subsection{Algorithm Pipelining and Configuration} \label{sec:av-pipelining}

\mast\ makes use of flexible python-based configuration files for automated building of sensor fusion pipelines. We take inspiration from OpenMMLab's configuration management~\cite{2020mmdet3d} and add algorithm registries to \mast\ and \avstack. Perception, tracking, clustering, and fusion algorithms are then configured in a high-level language such as in Configuration~\ref{lst:config}. Agents can run the same or different pipelines to suit the needs of the evaluation.

\begin{lstlisting}[language=Python,label={lst:config},caption=Pipelines configured with simple python code.]
# command_center_pipeline.py
commandcenter = dict(
    type="CommandCenter",
    pipeline = dict(
        type="CommandCenterPipeline",
        clustering=dict(
            type="SampledAssignmentClusterer",
            assign_radius=2.0),
        group_tracking=dict(
            type="GroupTrackerWrapper",
            fusion=dict(
                type="CovarianceIntersectionFusion"),
            tracker=dict(type="KalmanBoxTracker3D")
)))

\end{lstlisting}




\section{Threat Model for Security Analysis} \label{sec:4-threat-model}

\begin{figure}
    \centering
    \begin{subfigure}[b]{0.95\linewidth}
    \centering
        \includegraphics[width=0.72\linewidth]{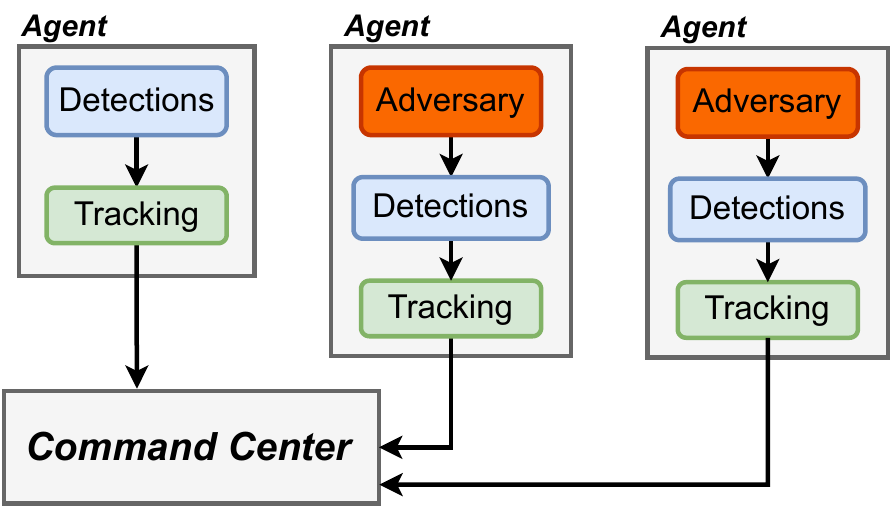}
        \caption{Uncoordinated adversaries.}
    \end{subfigure}
    \begin{subfigure}[b]{0.95\linewidth}
    \centering
        \includegraphics[width=0.9\linewidth]{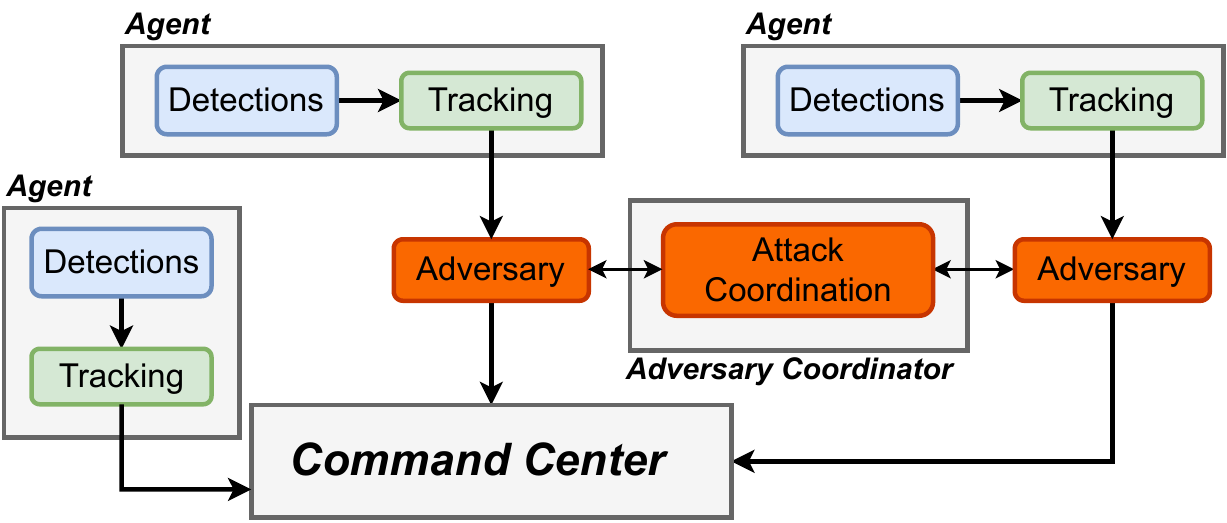}
        \caption{Coordinated adversaries.}
    \end{subfigure}
    \caption{(a) We model uncoordinated adversaries as directly manipulating detections from perception. Each adversary has its own local objective. This model can encompass physical access, sensing channels, and compromised model attack surfaces. (b) In the coordinated case, adversaries communicate with an attack coordinator to synchronize attack objective functions. We model this threat at the network level where communication links are compromised.}
    \label{fig:attack-diagram}
\end{figure}

We present and evaluate two threat models in this work: (1) a subset of agents are independently compromised at the sensing or perception level in an ``uncoordinated attack'', and (2) a subset of agents are compromised at the tracking or communication level by colluding adversaries in a ``coordinated attack''. The diagram in Fig.~\ref{fig:attack-diagram} illustrates graphically the positioning of the adversary in each case.

\myparagraph{Uncoordinated Attack} \label{sec:4-threat-model-uncoordinated}
Many existing security analyses focus on attacks at the sensing or perception level. One surface for this class of attack is to obtain physical access to the vehicle's electronic components. Physical access was used in recent CPS security case studies~\cite{bono2005security, 2010koschercheckoway, kamkar2015drive, ibrahim2019key, checkoway2011comprehensive}. Another attack surface is to infiltrate the physical sensing channels via black-box~\cite{2020sun-spoofing} or white-box spoofing attacks~\cite{2019cao-spoofing} or physical adversaries objects~\cite{2020tuphysicalatt, abdelfattah2021towards}. These attacks are ``uncoordinated'' - i.e., an adversary compromises sensing/detections of an AV without an ability to manipulate other AVs or communicate with other attacker nodes.

In an uncoordinated attack, each adversary is operating under its own malicious objective function. While the goal of the attackers may align (e.g.,~add fake objects), the implementation may be inconsistent across attackers (e.g.,~the location and/or type of attacks may differ).

We choose to model uncoordinated attacks as the attacker directly adding/removing detections at the output of perception. This model can account for physical access, sensing channels, and compromised perception attack surfaces.

\myparagraph{Coordinated Attack} \label{sec:4-threat-model-coordinated}
Communication networks are needed in multi-agent contexts to distribute information among agents. Connecting potentially untrusted agents, infrastructure sensing, and edge servers presents a new class of threat models for AVs: \textit{coordinated attacks}. Adversaries positioned in the communication network can leverage the network itself to synchronize attacks. Coordinated attacks are most likely to occur at the networking-level of a distributed system and can be realized as compromised Wi-Fi hotspots~\cite{malik2020analysis} or access points~\cite{benko2019security,petrillo2020secure}. 

A coordinated attack has all adversaries operating under a common objective function with pooled knowledge. The adversaries thus have their own ``command-center'' to coordinate the actions between them. For this reason, coordinated attacks are more powerful than uncoordinated attacks.

We choose to model coordinated attacks with the presence of an attack coordinator node. Adversaries send agent pose and tracked objects to the coordinator. It is the coordinator's job to decide on the false positive/negative targets and to share this information with each of the adversaries. The adversaries then apply this attack directive to their compromised agent.



\subsection{Attack Knowledge and Capability}

\textit{Uncoordinated attacks.} \,
We assume an adversary that has compromised an agent has full knowledge of the agent's pose and perception data. We model the attacker as being able to directly modify the detections. This encapsulates the effects of manipulating physical channels (e.g.,~spoofing~\cite{2020sun-spoofing,2022hally-frustum}, adversarial objects~\cite{2020tuphysicalatt,abdelfattah2021towards}) or cyber-based attackers~\cite{hallyburton2023securing}. 

\textit{Coordinated attacks.} \,
A coordinated attack exists at the network level. As a result, attackers have access to both the agent's pose and the agent's estimated state of objects in the scene. These estimated states are in the form of object tracks with at least position, bounding-box, velocity, and orientation states. An attacker manipulates the estimated states of surrounding objects by adding false positives, removing tracks to create false negatives, or perturbing the state of an existing track to create a translation, as defined in~\cite{2022hally-frustum}. 

\subsection{Attacker Goals} 

The goal of the attackers in both threat models is to corrupt the situational awareness of the \commandcenter. This goal is impactful because of the redistribution of information to nearby agents; information flows from the \commandcenter\ back to all agents \textit{including those agents who were not directly compromised by an adversary}, as described in Fig.~\ref{fig:multi-agent-pipeline}. Distributing compromised information to agents interferes with mission-critical functions. We consider two objective attack goals in this work: the attacker wishes to (1) add fake (spoofed) objects (false positives), and (2) remove existing objects (false negatives) at the \commandcenter.

\subsection{Attack Designs}

The attack's power is a function of the number of agents that are compromised and of the number of manipulations performed (i.e., number of false positives added, true objects negated). A more powerful attack has more compromised agents and/or performs more manipulations. Table~\ref{tab:security-params} describes the design parameters evaluated in these security studies.

\begin{table}[t]
    \centering
    \resizebox{\linewidth}{!}{%
    \begin{tabular}{c|c}
        \hline
        Parameter & Type \\
        \toprule
        Num. of compromised agents & Int. on [0, \# agents] \\
        Pct. of existing detections manipulated & Pct on [0, 1.0] \\
        Num. of false detections added & Positive integer \\
        (Un)Coordinated & Boolean \\
        \bottomrule
    \end{tabular}}
    \caption{Security analysis varies adversary capability including number of compromised nodes and level of data corruption.}
    \label{tab:security-params}
\end{table}

\subsection{Attack Effectiveness}

We evaluate attack effectiveness along two dimensions: (1) The attacker must remain stealthy to integrity algorithms from the \commandcenter. (2) The attacker is effective if he causes fake objects at the \commandcenter\ or if true objects in an agent's field of view are not detected at the \commandcenter. To quantify attack success, we derive metrics to describe the agent/\commandcenter\ situational awareness compared to ground truth object locations. 

We only attack the infrastructure sensors in this work and assume that situational awareness local to the ego vehicle is free from adversarial manipulation. This choice was made to focus our security analysis. An increased emphasis on securing autonomous vehicles and the high degree of difficulty in exercising physical attacks on mobile AVs suggests that ownship information could be secured. Errors including false positives, false negatives, and noisy readings can still occur. 
\section{Attack Executions}

We present the attack executions. The attacker's goal is to compromise the situational awareness at the \commandcenter\ while remaining stealthy to integrity algorithms. Attackers begin by selecting targets. Once targets are selected, the adversaries will manipulate data to achieve their target outcomes.

\subsection{Target Object Selection }
The two threat models differ in how they select false positive and false negative targets. The uncoordinated attack selects targets based purely on local information while in the coordinated attack, an adversary coordinator directs the targets. 

\myparagraph{Uncoordinated Attacks} To select targets, each adversary uses the compromised agent's detected objects as input to its own instance of the selection algorithm. Target false negatives are selected as some fraction of objects from an initial set of the agent's detections. False positives are chosen as pseudo-random states within a maximum distance from the agent.

\myparagraph{Coordinated Attacks} Target selection occurs at the adversary coordinator. Each adversary sends uncompromised tracks from their host agent to the coordinator. The coordinator aligns the time and frame of all tracks and follows the selection algorithm of choosing target false negatives from clusters of objects and false positives as pseudo-random states close to any existing agent. After target selection, the coordinator sends the targets back to the adversaries who apply the directive.

\subsection{Data Manipulation}
The general framework of instantiating the attacks is identical in both the uncoordinated and coordinated cases given a selection of targets. After target selection, the adversary determines which detections/tracks to negate by performing an assignment between the agent's detections/tracks and the target false negatives. Detections/tracks matching a false negative target are removed. False positive targets are propagated to the current time and appended to the detection/track list if they are within a maximum distance from the agent.
\section{Evaluation} \label{sec:6-evaluation}

\mast\ is a useful tool for case studies and Monte Carlo evaluations. We present case studies and visualizations from (un)coordinated attacks against multi-agent collaborative fusion. We then present a Monte Carlo analysis over scenes and adversary configurations. In all cases, we fix the agent and \commandcenter\ pipelines as described in Section~\ref{sec:2-multi-agent-avs} for the simplicity of comparison. Fixing pipelines is not a requirement as \mast\ can easily spin up different pipelines using high-level configuration files. Table~\ref{tab:visualization-key} provides the color map key for visualizations.

\begin{table}[t]
    \centering
    \begin{tabular}{p{0.13\linewidth} | p{0.75\linewidth}}
        \textbf{Box Color} & \textbf{Indication} \\
        \toprule
        White & The ground-truth state of an object with significant overlap with a detection/track. Always pairs with blue. ``True positive''.\\
        Blue & The detection/track state of an object with significant overlap with a ground-truth state. Always pairs with white. ``True positive''.\\
        Yellow & The ground-truth state of an object lacking overlap with a detection/track. ``False negative''.\\
        Red & The detection/track state of an object lacking overlap with a ground-truth state. ``False positive''.\\
        Green & Object state (any of ground truth, detection, or track) when no attempt at assignment/evaluation is made. For visualization only.\\
        \bottomrule
    \end{tabular}
    \caption{Case study visualizations color key.}
    \label{tab:visualization-key}
\end{table}

\subsection{Case Studies}

All case studies use scene \#1 from the CARLA multi-agent dataset~\cite{hallyburton2023datasets}. Four infrastructure sensors capture camera and \lidar\ data from elevated positions near the ego vehicle.

The target selection of the uncoordinated adversaries and of the adversary coordinator follow the same \textit{general} framework. In both cases, the target selector samples a number of false positives via $k_{FP} \sim \text{Pois}(\lambda)$ and a number of false negatives via $k_{FN} \sim r \cdot |D|$ where $\lambda$ is the Poisson distribution parameter, $r$ is a percentage, and $|D|$ is the number of objects detected by the agent in the uncoordinated case and the number of clusters in the coordinated case. We fix $\lambda=5\ \text{and}\ r=20\%$ for all case studies. Fixing the parameters to be identical for both threat models means that more manipulations will occur in the uncoordinated case because there are a number of target selections happening equal to the number of adversaries compared to a single target selection for the coordinated case.

\subsubsection{Baseline}
Natural occlusions between objects and limited sensor range will truncate the ego's understanding of the environment. This will materialize as a number of false negatives from the ego's local tracks, as seen in Figs.~\ref{fig:case-study-baseline-a},~\ref{fig:case-study-baseline-c}. The centralized fusion of infrastructure sensing data can mitigate the incidence of false negatives by providing complementary perspectives. This benefit of multi-agent fusion is on display in Figs.~\ref{fig:case-study-baseline-b},~\ref{fig:case-study-baseline-d}. Ego and infrastructure tracks are fused at the \commandcenter\ and redistributed to the ego for mission-critical planning. Multi-agent fusion has eliminated all false negatives from the ego's situational awareness without introducing any false positives.

\providecommand\imgtrimm{}
\providecommand\imgrtrimm{}
\renewcommand{\imgtrimm}{10}
\renewcommand{\imgrtrimm}{14}

\providecommand\lidltrimm{}
\providecommand\lidbtrimm{}
\providecommand\lidrtrimm{}
\providecommand\lidttrimm{}
\renewcommand{\lidltrimm}{5}
\renewcommand{\lidbtrimm}{7}
\renewcommand{\lidrtrimm}{2}
\renewcommand{\lidttrimm}{1}

\begin{figure}
    \centering
    \begin{subfigure}[b]{0.95\linewidth}
        \centering
        \fbox{\includegraphics[width=0.9\linewidth,trim={\imgtrimm cm \imgtrimm cm \imgrtrimm cm \imgtrimm cm},clip]{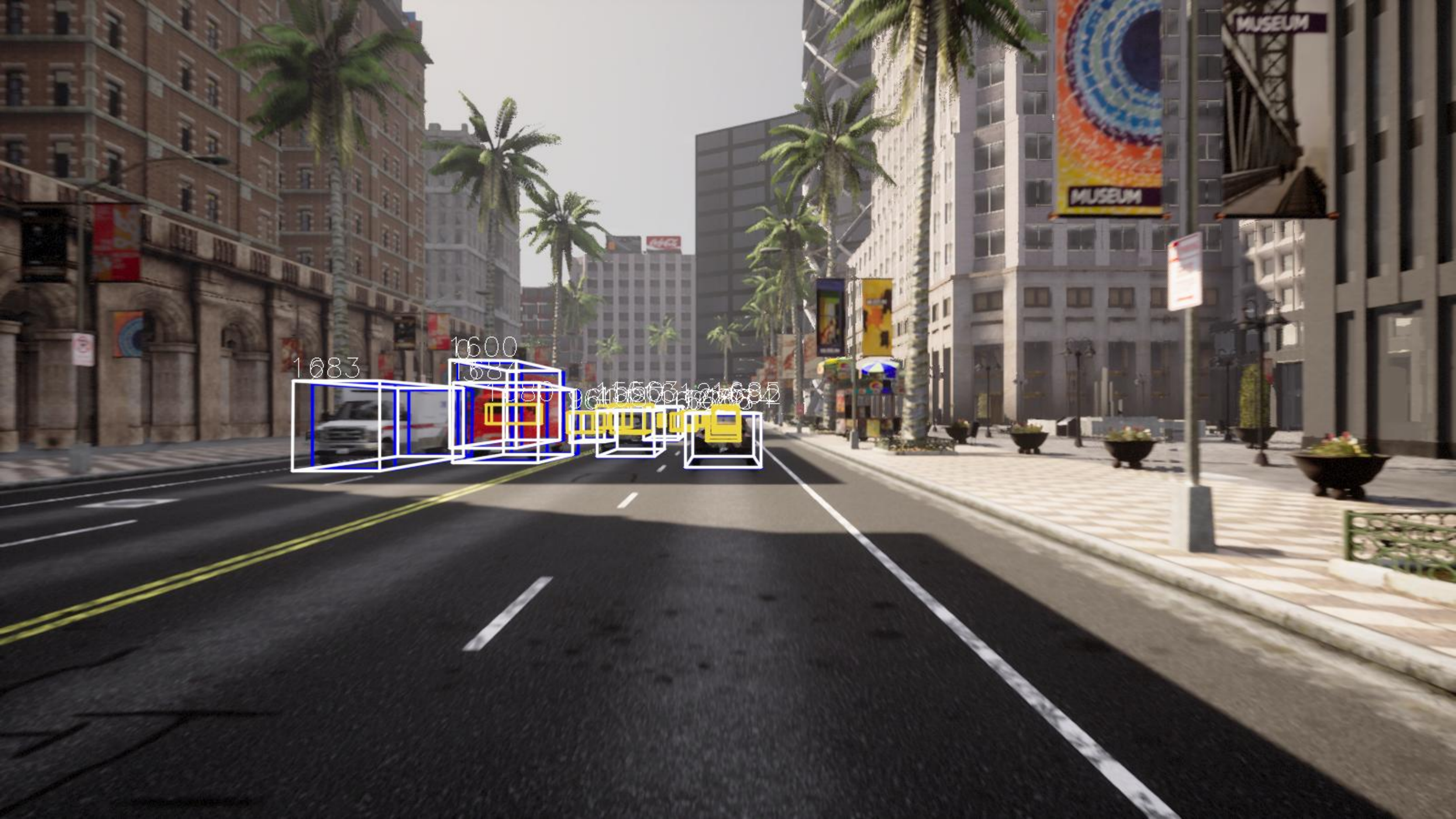}}
        \caption{\textit{Baseline:} ego view with local tracks. False negatives (yellow) occur because ego suffers from occlusion.}
        \label{fig:case-study-baseline-a}
    \end{subfigure}
    \begin{subfigure}[b]{0.95\linewidth}
        \centering
        \fbox{\includegraphics[width=0.9\linewidth,trim={\imgtrimm cm \imgtrimm cm \imgrtrimm cm \imgtrimm cm},clip]{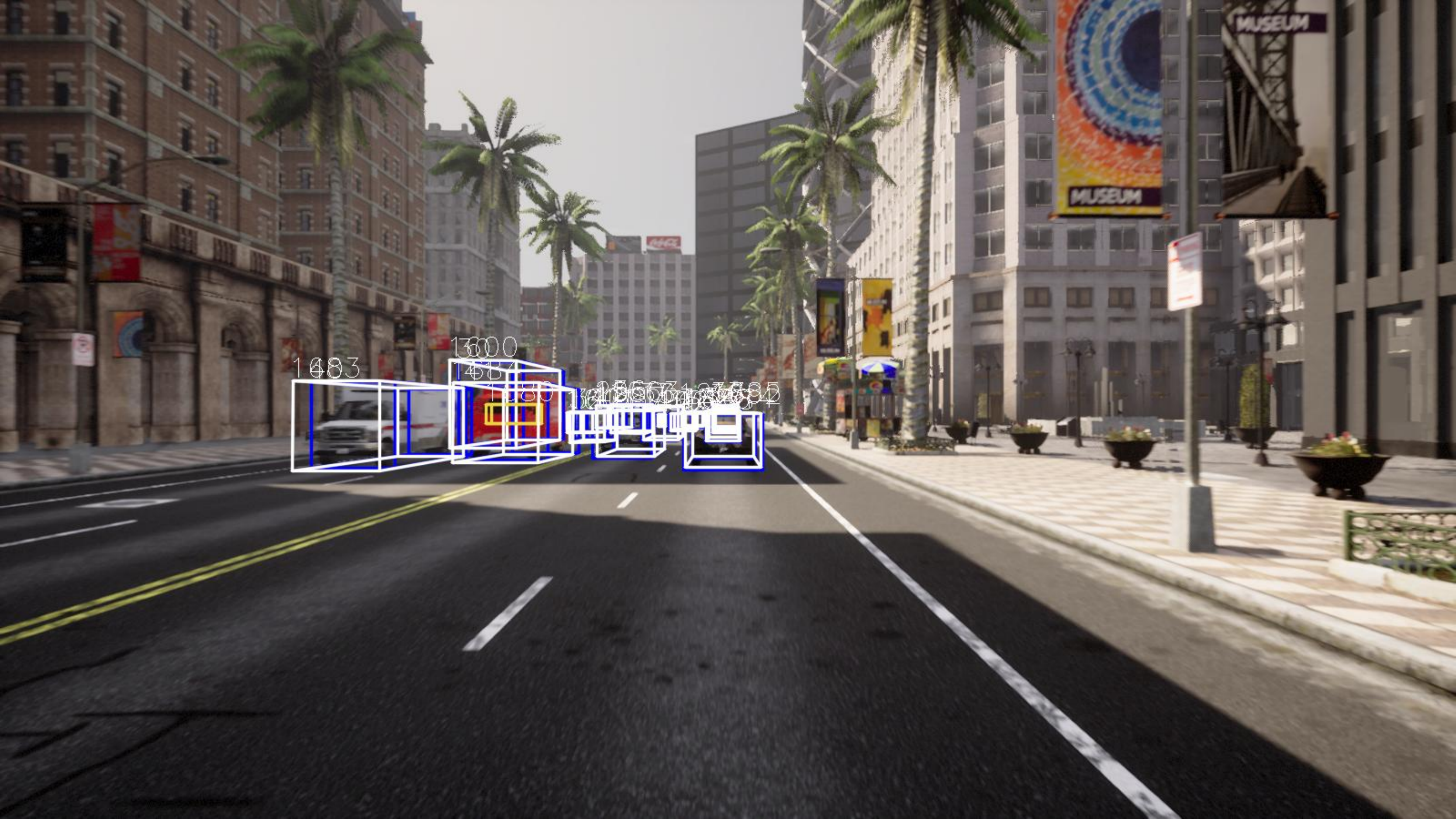}}
        \caption{\textit{Baseline:} ego view with fused tracks from \commandcenter. Occlusion mitigated; more objects detected at a longer range.}
        \label{fig:case-study-baseline-b}
    \end{subfigure}
    \begin{subfigure}[b]{0.45\linewidth}
        \centering
        \fbox{\includegraphics[width=0.9\linewidth,trim={\lidltrimm cm \lidbtrimm cm \lidrtrimm cm \lidttrimm cm},clip]{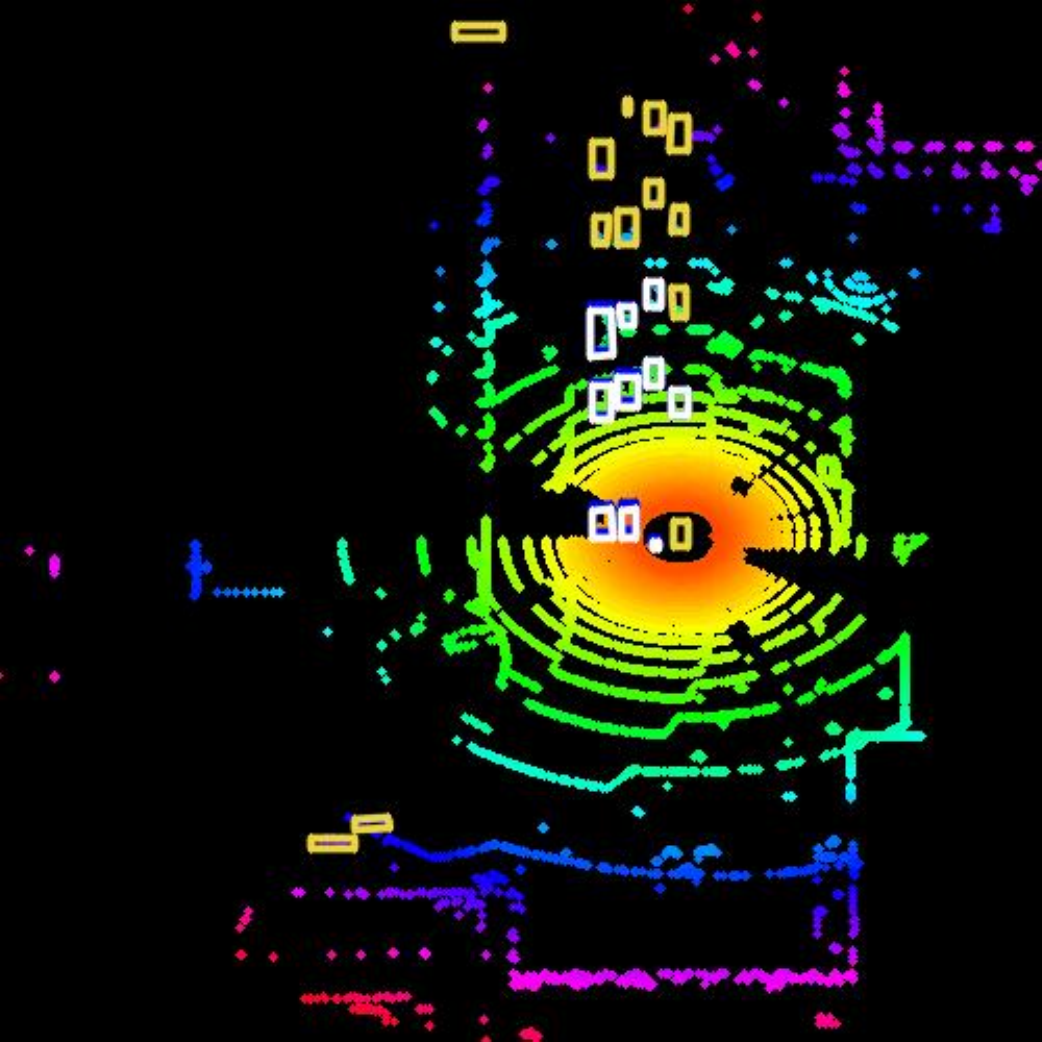}}
        \caption{Companion of (a); BEV of ego LiDAR with local information illustrates large number of false negatives in yellow.}
        \label{fig:case-study-baseline-c}
    \end{subfigure}
    \begin{subfigure}[b]{0.45\linewidth}
        \centering
        \fbox{\includegraphics[width=0.9\linewidth,trim={\lidltrimm cm \lidbtrimm cm \lidrtrimm cm \lidttrimm cm},clip]{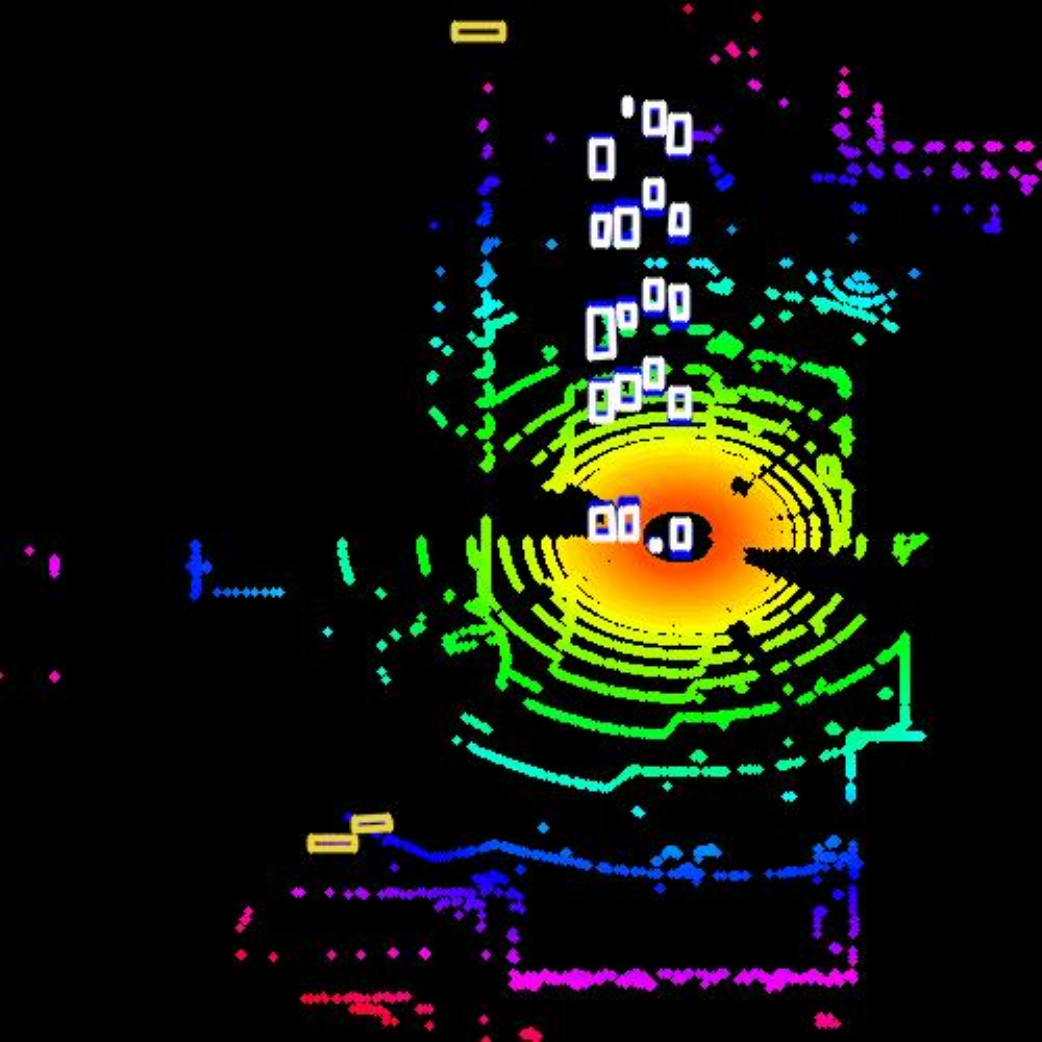}}
        \caption{Companion of (b); BEV of ego LiDAR with uncompromised \commandcenter\ information mitigates occlusion.}
        \label{fig:case-study-baseline-d}
    \end{subfigure}
    \begin{subfigure}[b]{0.95\linewidth}
        \centering
        \fbox{\includegraphics[width=0.9\linewidth,trim={
        2cm 4cm 2cm 5cm
        },clip]{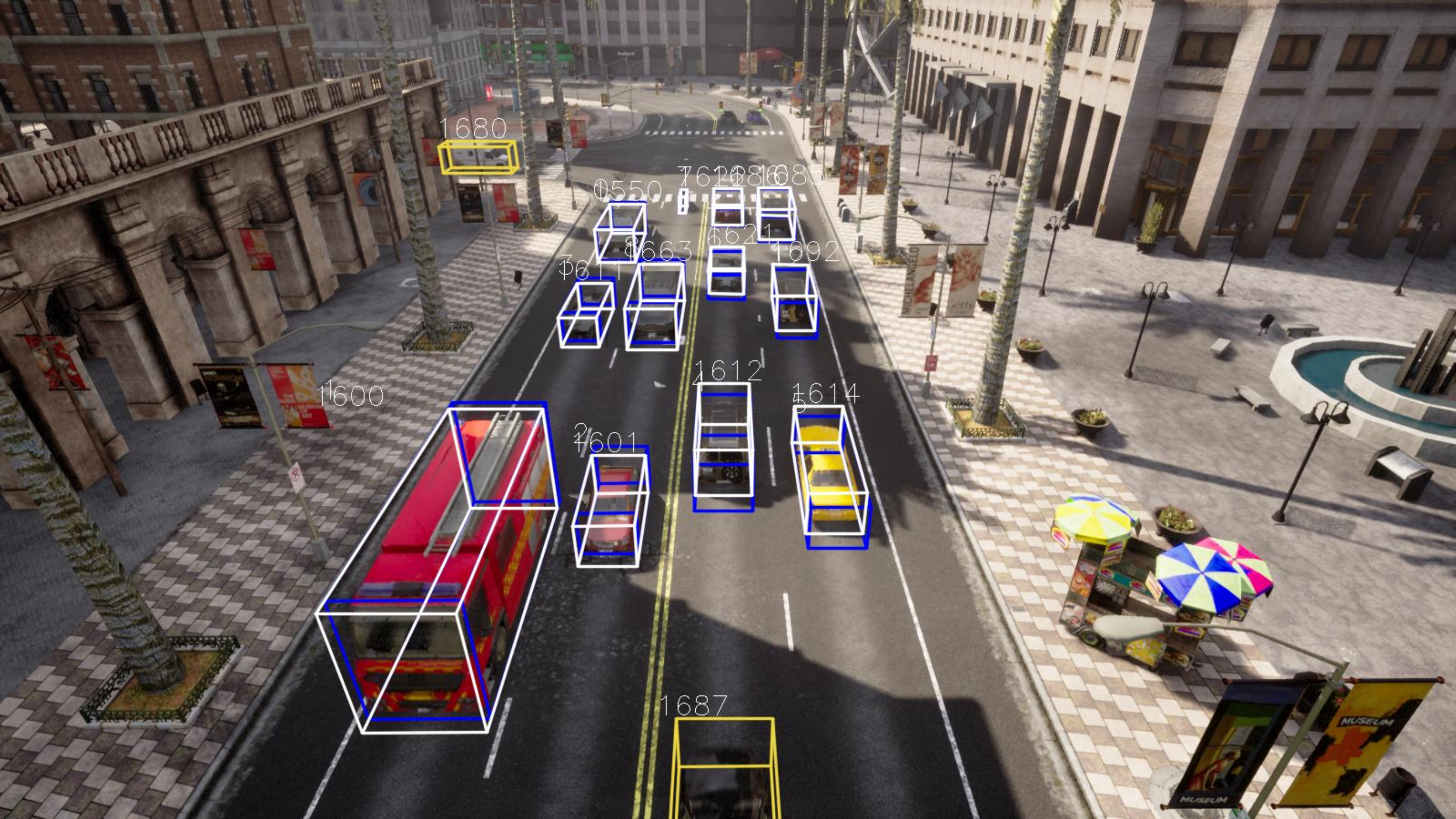}}
        \caption{Uncompromised infrastructure agent with local track states.}
    \end{subfigure}
    \caption{\textit{Baseline:} (a) Ego experiences strong occlusion in local information compared to (b) fusing multi-agent information from \commandcenter. (c, d) Bird's-eye view (BEV) companions to (a, b) with boxes on \lidar\ data. (e) An infrastructure sensor provides a collaborative vantage point on the same scene. Multiple agents can mitigate ground-vehicle occlusions.}
    \label{fig:case-study-baseline}
\end{figure}

\subsubsection{Uncoordinated Attack}
In an uncoordinated attack, adversaries select targets independent of each other. Following the threat model in Section~\ref{sec:4-threat-model-uncoordinated}, we simply model uncoordinated attacks as compromised detections at the output of perception.

Two adversaries latch on to two of the four infrastructure sensors. Each adversary compromises the agent locally without regard for the other. This creates false positives/negatives for the agents, as in Figs.~\ref{fig:case-study-uncoord-a},~\ref{fig:case-study-uncoord-c}. The compromised agents send attacked state estimates to the \commandcenter. Without integrity to detect the attacks, the \commandcenter\ will be partially compromised. As information flows back from the \commandcenter\ to the agents, a compromised \commandcenter\ will introduce false positives/negatives into the ego's fused state. In Figs.~\ref{fig:case-study-uncoord-b},~\ref{fig:case-study-uncoord-d}, the ego's fused state sees many false positives compared to the baseline of Fig.~\ref{fig:case-study-baseline}.

\providecommand\imgtrimm{}
\providecommand\imgrtrimm{}
\providecommand\agentimgltrimm{}
\providecommand\agentimgbtrimm{}
\providecommand\agentimgrtrimm{}
\providecommand\agentimgttrimm{}
\renewcommand{\imgtrimm}{6}
\renewcommand{\imgrtrimm}{12}
\renewcommand{\agentimgltrimm}{2}
\renewcommand{\agentimgbtrimm}{0}
\renewcommand{\agentimgrtrimm}{2}
\renewcommand{\agentimgttrimm}{7}

\providecommand\lidltrimm{}
\providecommand\lidbtrimm{}
\providecommand\lidrtrimm{}
\providecommand\lidttrimm{}
\renewcommand{\lidltrimm}{5}
\renewcommand{\lidbtrimm}{5}
\renewcommand{\lidrtrimm}{2}
\renewcommand{\lidttrimm}{1}

\begin{figure}
    \begin{subfigure}[b]{0.95\linewidth}
        \centering
        \fbox{\includegraphics[width=0.9\linewidth,trim={
            \agentimgltrimm cm \agentimgbtrimm cm \agentimgrtrimm cm \agentimgttrimm cm
        },
        clip]{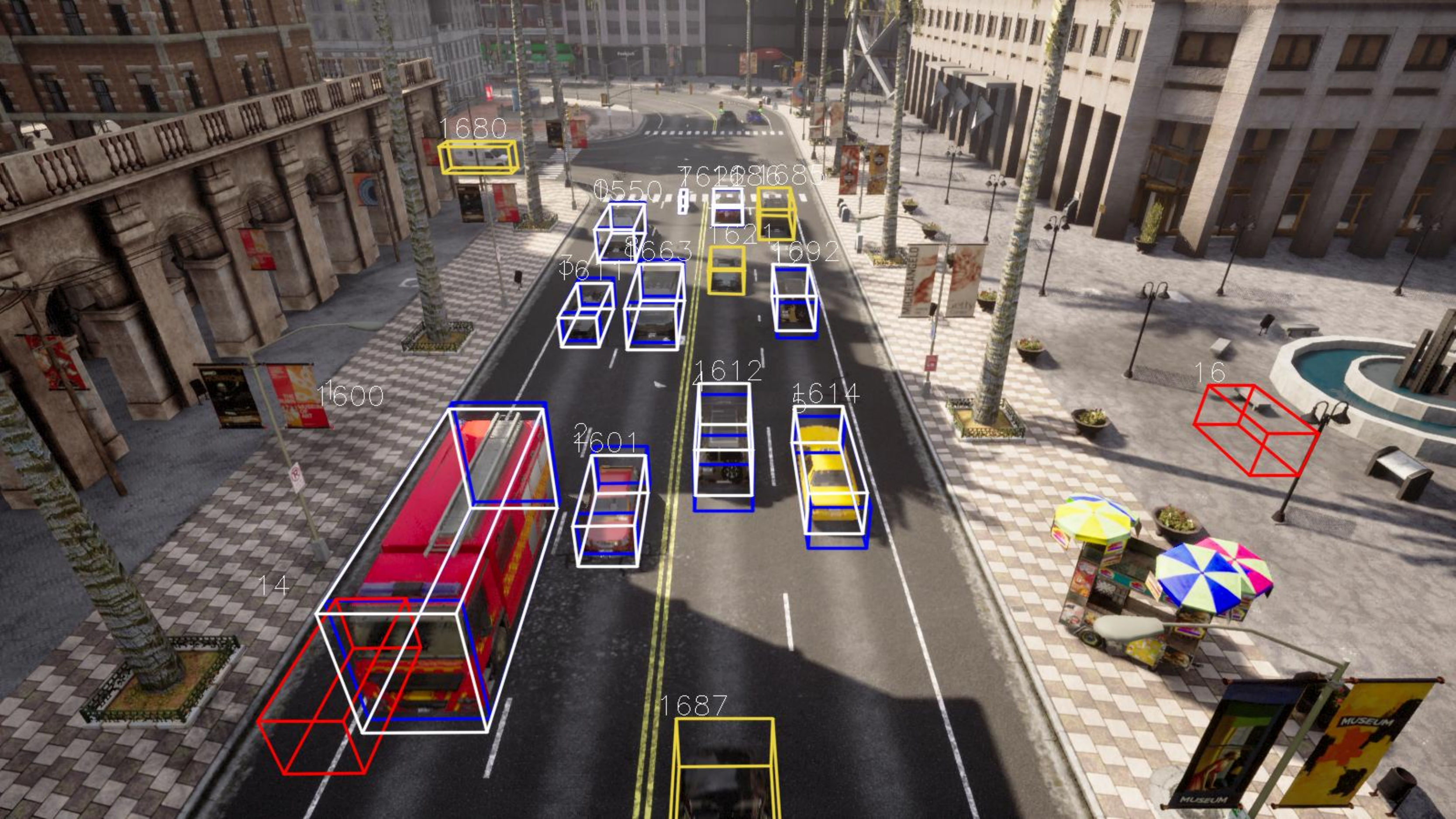}}
        \caption{\textit{Uncoord:} agent \#1 (attacked) with local tracks. False negatives (yellow) and positives (red) occur naturally \textit{and} due to adversary.}
        \label{fig:case-study-uncoord-a}
    \end{subfigure}
    %
    %
    \centering
    \begin{subfigure}[b]{0.95\linewidth}
        \centering
        \fbox{\includegraphics[width=0.9\linewidth,trim={
            \imgtrimm cm \imgtrimm cm \imgrtrimm cm \imgtrimm cm
        },clip]{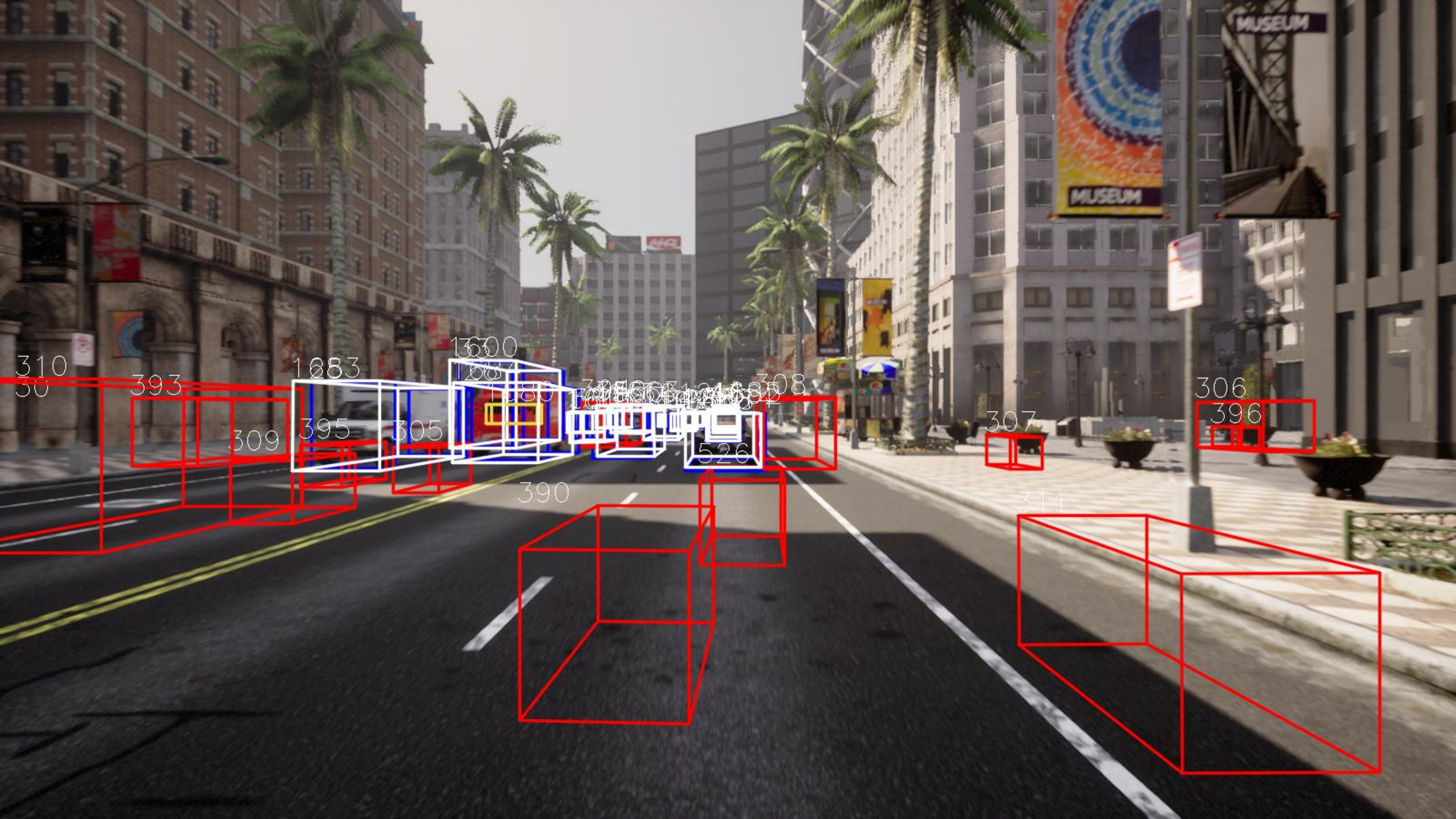}}
        \caption{\textit{Uncoord:} unattacked ego with fused tracks from \commandcenter. Attacked agents have compromised \commandcenter; many false positives sent to ego.}
        \label{fig:case-study-uncoord-b}
    \end{subfigure}
    %
    \begin{subfigure}[b]{0.45\linewidth}
        \centering
        \fbox{\includegraphics[width=0.9\linewidth,trim={
            \lidltrimm cm \lidbtrimm cm \lidrtrimm cm \lidttrimm cm
        },clip]{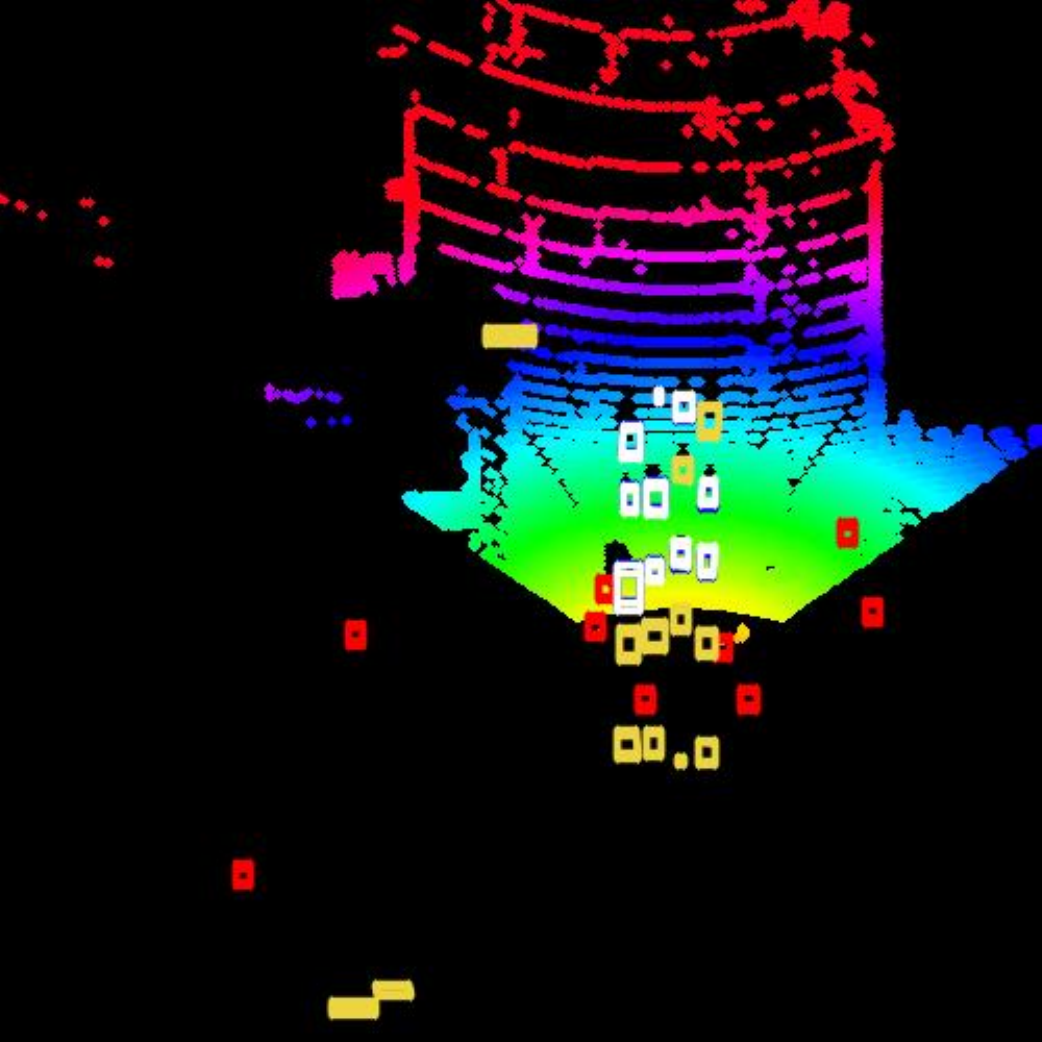}}
        \caption{Companion of (a); BEV of agent \lidar\ with compromised local information.}
        \label{fig:case-study-uncoord-c}
    \end{subfigure}
    \begin{subfigure}[b]{0.45\linewidth}
        \centering
        \fbox{\includegraphics[width=0.9\linewidth,trim={
            \lidltrimm cm \lidbtrimm cm \lidrtrimm cm \lidttrimm cm
        },clip]{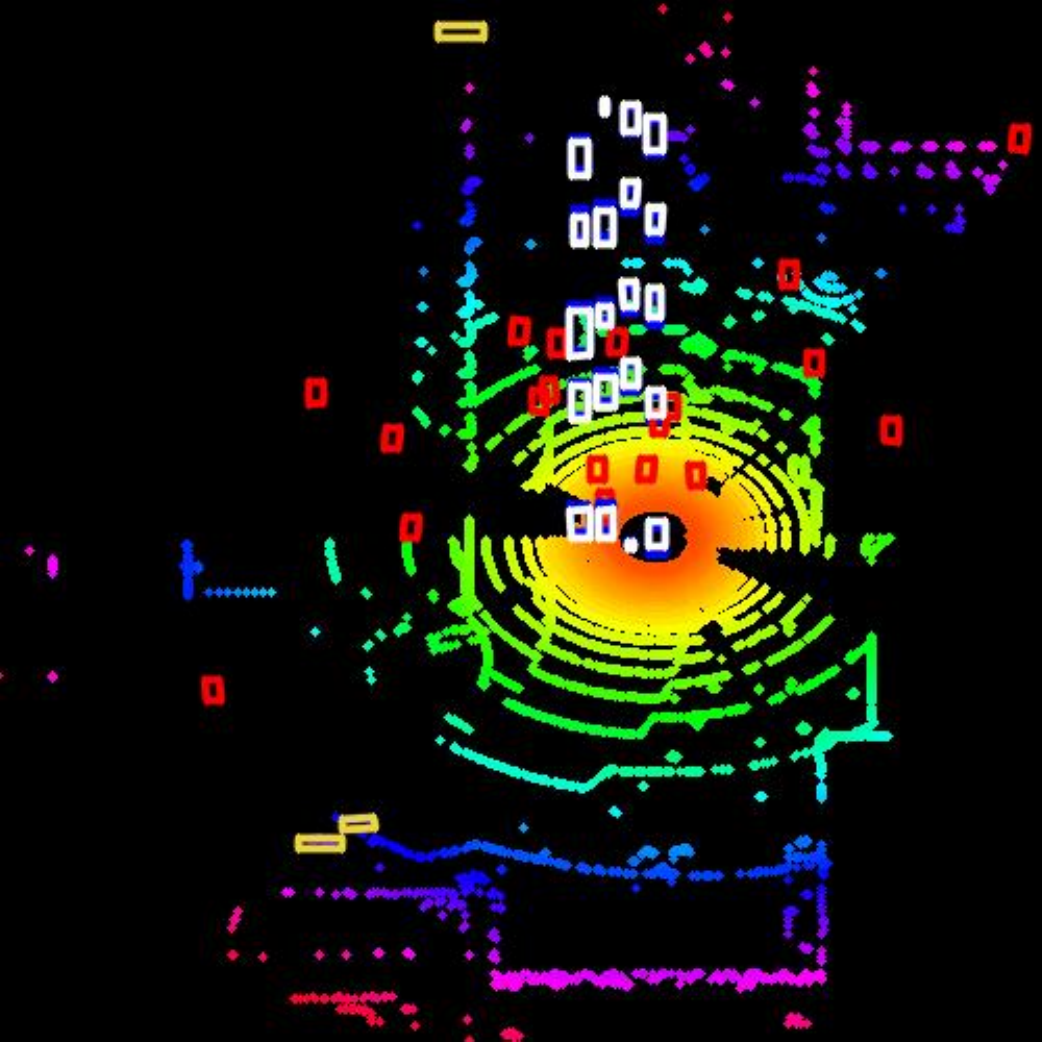}}
        \caption{Companion of (b); BEV of ego \lidar\ with compromised \commandcenter\ information.}
        \label{fig:case-study-uncoord-d}
    \end{subfigure}
    \caption{Two of four agents are compromised in uncoordinated attack. (a) Compromised agent has attacker-induced false positives and negatives. (b) Attacks compromise \commandcenter\ and propagate into ego's fused state. Ego does not suffer from false negatives because other agents can correctly identify existing objects. This is due to attacks being uncoordinated. (c, d) BEV companions to (a, b) with track states marked in boxes.}
    \label{fig:case-study-uncoord}
\end{figure}

In the uncoordinated case, there is low likelihood that two adversaries will have consistent targets. The adversaries can still bring about false positives at the \commandcenter\ even though the attacks are uncoordinated. Without integrity, the \commandcenter\ is incapable of filtering out false tracks from individual agents if those false tracks are self-consistent longitudinally. However, there will be few false negatives at the \commandcenter\ because the same object is unlikely to be negated by multiple adversaries.

\subsubsection{Coordinated Attacks}
In a coordinated attack, the attack coordinator selects global targets and passes consistent attack directives back to each of the adversaries. Following the threat model of Section~\ref{sec:4-threat-model-coordinated}, we model coordinated attacks as compromised tracks in the multi-agent communication network. Such attacks are possible given the rise in connected vehicles and the sophistication of networking attacks~\cite{nhtsa2016}.

The coordinated attack parameters are identical to the uncoordinated case with the additional attack coordinator. Fig.~\ref{fig:case-study-coord} illustrates the results of the coordinated attack. False positives and negatives occur in the coordinated attack. The number of false positives is relatively lower compared to the uncoordinated case because the parameters of the lone coordinator were set identical to the parameters of \textit{each} of the two uncoordinated adversaries. However, when integrity algorithms are introduced into the \commandcenter, we expect uncoordinated false positives to be filtered out while coordinated false positives will create persistent outcomes at the \commandcenter.

\providecommand\imgtrimm{}
\providecommand\imgrtrimm{}
\providecommand\agentimgltrimm{}
\providecommand\agentimgbtrimm{}
\providecommand\agentimgrtrimm{}
\providecommand\agentimgttrimm{}
\renewcommand{\imgtrimm}{6}
\renewcommand{\imgrtrimm}{12}
\renewcommand{\agentimgltrimm}{2}
\renewcommand{\agentimgbtrimm}{0}
\renewcommand{\agentimgrtrimm}{2}
\renewcommand{\agentimgttrimm}{7}

\providecommand\lidltrimm{}
\providecommand\lidbtrimm{}
\providecommand\lidrtrimm{}
\providecommand\lidttrimm{}
\renewcommand{\lidltrimm}{5}
\renewcommand{\lidbtrimm}{5}
\renewcommand{\lidrtrimm}{2}
\renewcommand{\lidttrimm}{1}

\begin{figure}
    \begin{subfigure}[b]{0.95\linewidth}
        \centering
        \fbox{\includegraphics[width=0.9\linewidth,trim={
            \agentimgltrimm cm \agentimgbtrimm cm \agentimgrtrimm cm \agentimgttrimm cm
        },
        clip]
        {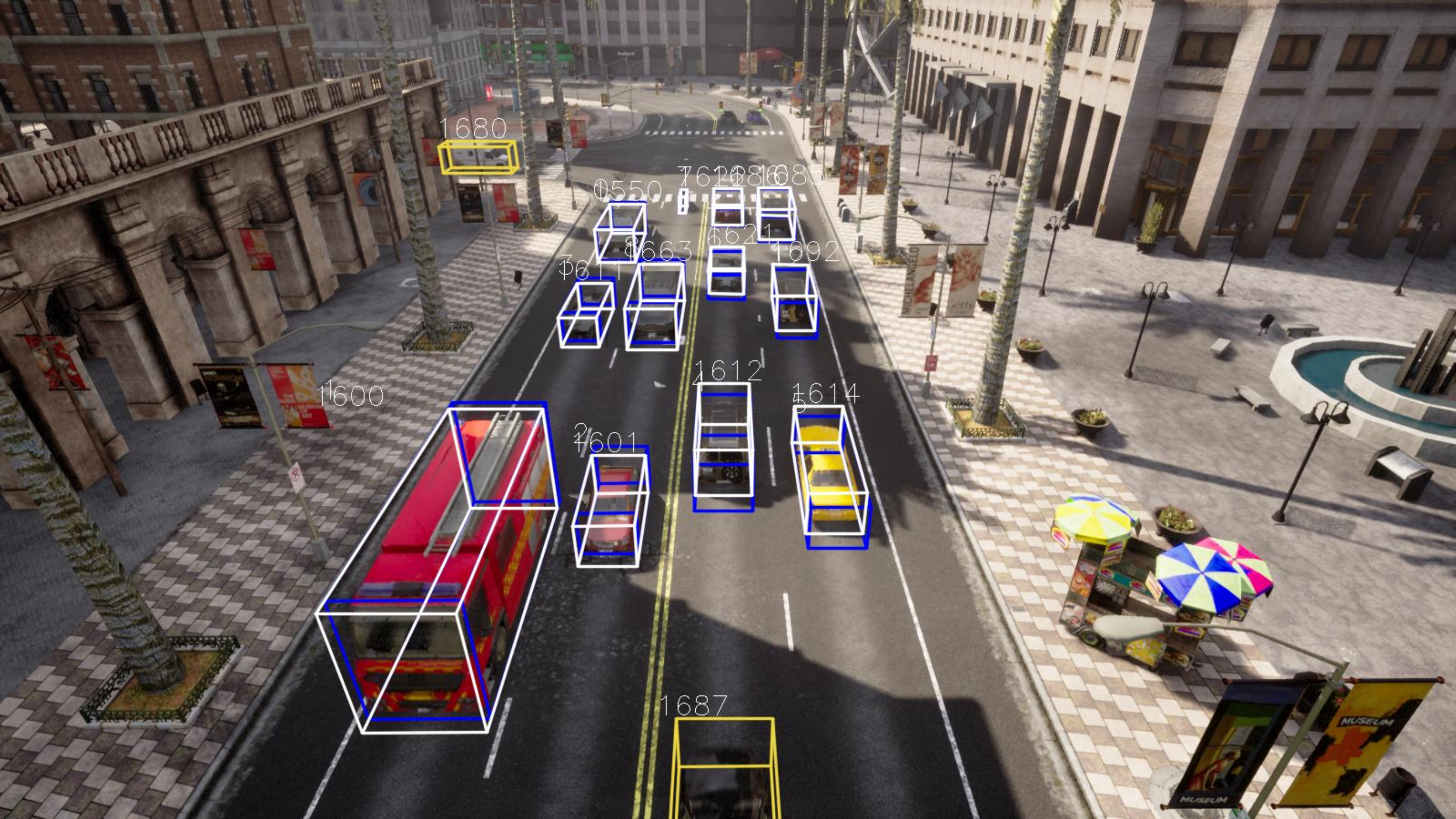}
        }
        \caption{\textit{Coord:} agent \#1 (attacked) with local tracks. Fewer false positives occur compared to uncoordinated due to attack parameters.}
    \end{subfigure}
    %
    %
    \centering
    \begin{subfigure}[b]{0.95\linewidth}
        \centering
        \fbox{\includegraphics[width=0.9\linewidth,trim={
            \imgtrimm cm \imgtrimm cm \imgrtrimm cm \imgtrimm cm
        },clip]
        {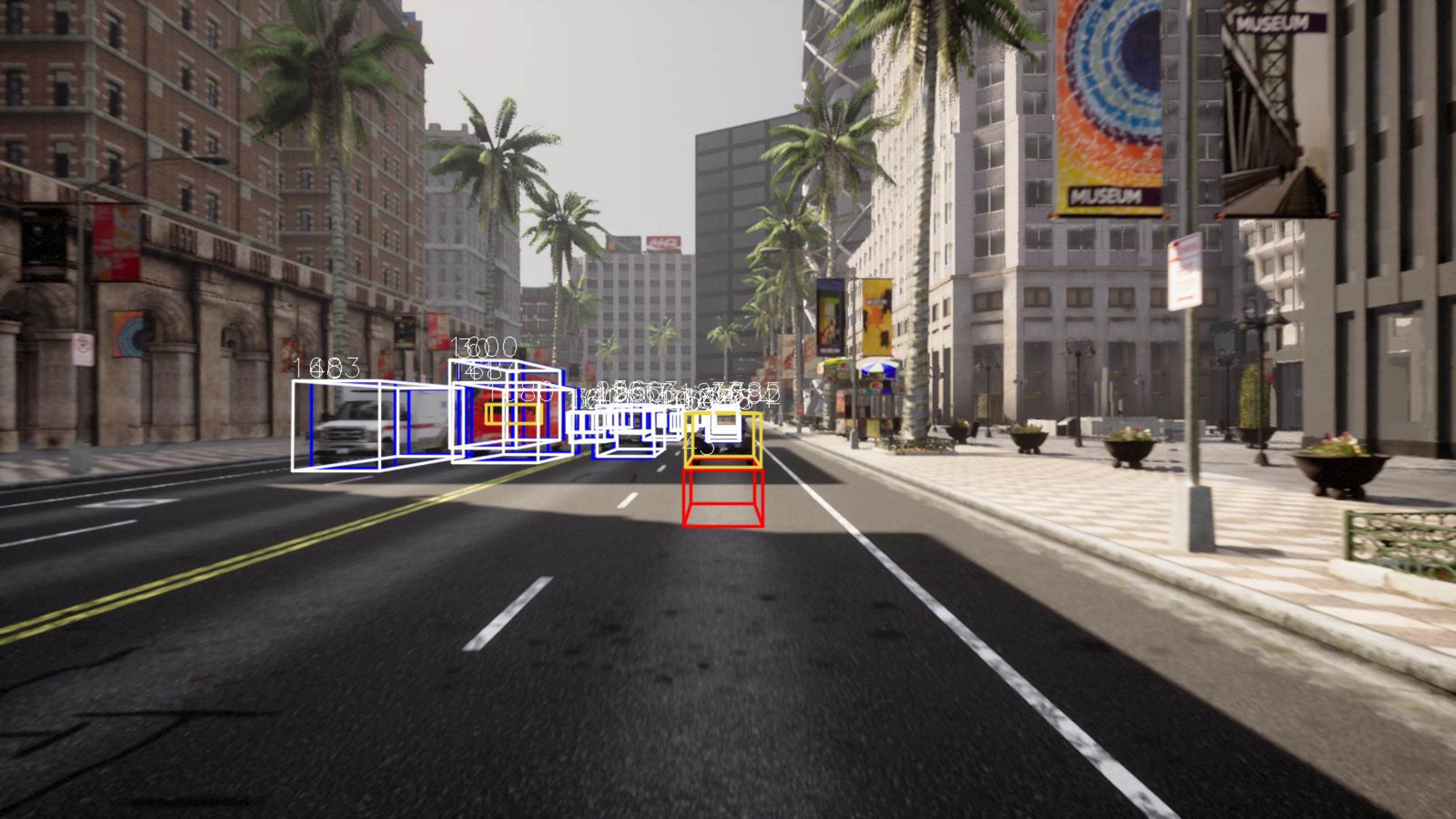}
        }
        \caption{textit{Coord:} ego view with fused tracks from \commandcenter. Attacked agents have compromised \commandcenter.}
    \end{subfigure}
    %
    \begin{subfigure}[b]{0.45\linewidth}
        \centering
        \fbox{\includegraphics[width=0.9\linewidth,trim={
            \lidltrimm cm \lidbtrimm cm \lidrtrimm cm \lidttrimm cm
        },clip]
        {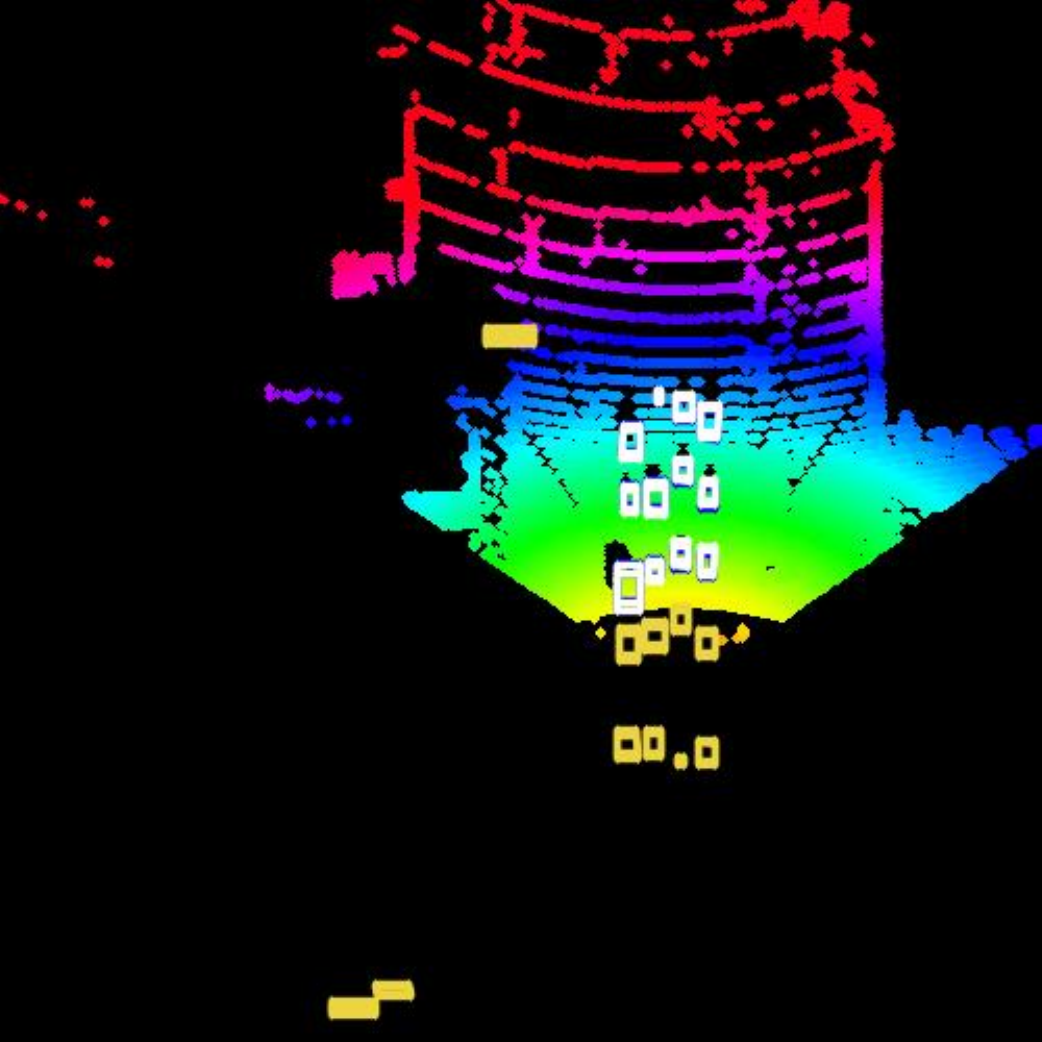}
        }
        \caption{Companion of (a); BEV of agent \lidar\ with compromised local information.}
    \end{subfigure}
    \begin{subfigure}[b]{0.45\linewidth}
        \centering
        \fbox{\includegraphics[width=0.9\linewidth,trim={
            \lidltrimm cm \lidbtrimm cm \lidrtrimm cm \lidttrimm cm
        },clip]
        {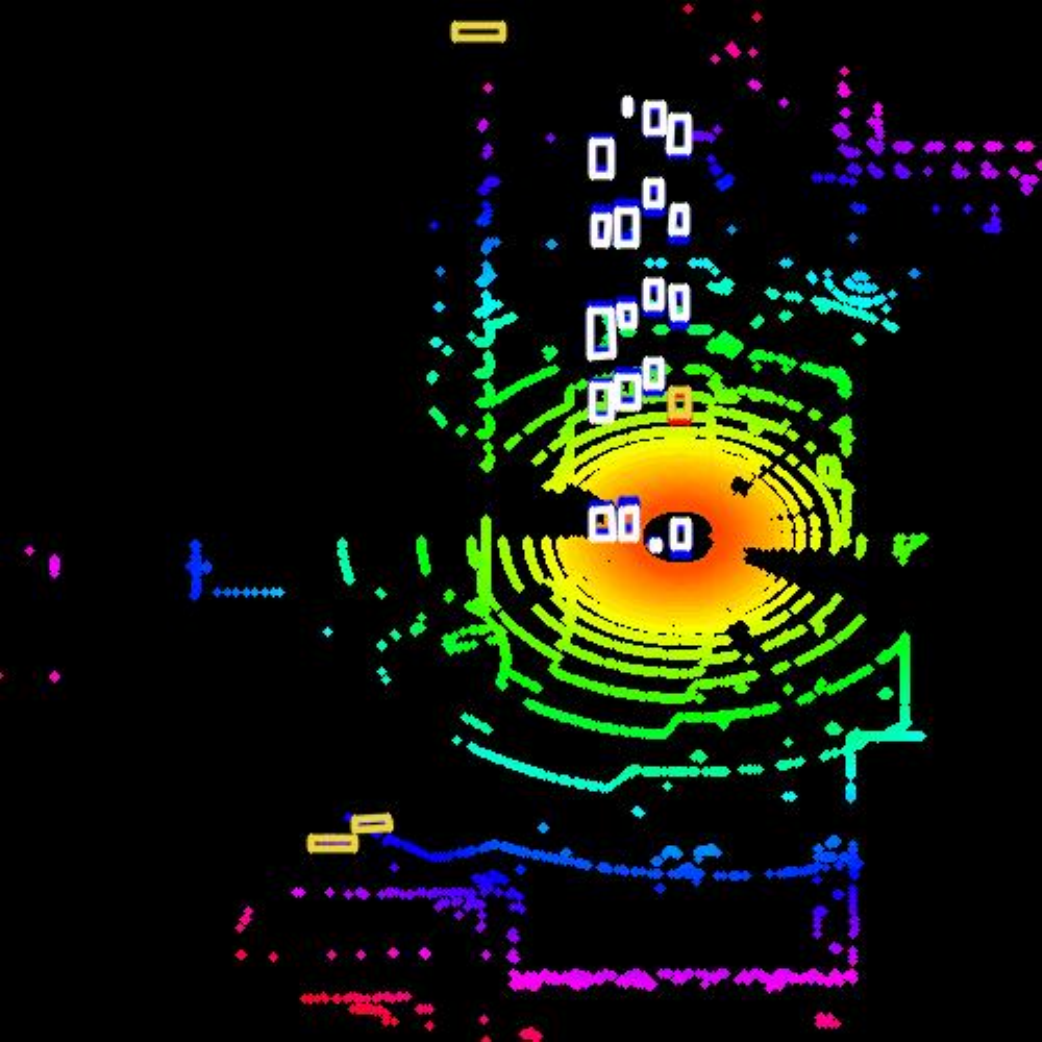}
        }
        \caption{Companion of (b); BEV of ego \lidar\ with compromised \commandcenter\ information.}
    \end{subfigure}
    \caption{Two of four agents are compromised in coordinated attack. Outcomes similar to uncoordinated case except that fewer false positives occur. This is a result of the attack parameters set for the experiments.}
    \label{fig:case-study-coord}
\end{figure}

\subsection{Monte Carlo Results} \label{sec:6-evaluation-mc}

We run Monte Carlo (MC) analysis over multiple scenes and attacker parameters to illustrate \mast's effectiveness for \msma\ security analysis. We vary the following parameters and hypothesize their effects. \\
\noindent \textbf{MC Parameters:}
\begin{itemize}
    \item \textbf{Attack type:} Coordinated vs. uncoordinated. Uncoordinated may generate more false positives due to parameters and lack of integrity at \commandcenter. Coordinated adversaries will be more successful obtaining false negatives.
    \item \textbf{Number of adversaries:} In the uncoordinated case without integrity at the \commandcenter, more adversaries leads to more false positives. More adversaries enable coordinated adversaries to overcome \commandcenter\ integrity algorithms when present in future works.
\end{itemize}

\subsubsection{Metrics}
The attackers' goals are to deteriorate the \commandcenter's situational awareness by effecting changes to the agents' local fusion. To quantify the outcomes, we run baseline and attack scenarios and determine performance of each agent relative to ground truth. The set of ground truth objects is made relative to each agent and constructed as objects in the scene within a threshold distance to that agent that are visible from at least one of any of the agents (mobile or static). Given this set of truth, we compute false positive and false negative outcomes for local and fused information. We capture metrics compared to truth both for agents' local information and after the \commandcenter\ has redistributed aggregated tracks to each agent.

We capture the following metrics in the analysis. \\
\noindent \textbf{MC Metrics:}
\begin{itemize}
    \item \textbf{CAFP(FN)IoB:} Compromised agent false positive/negative increment over baseline.
    \item \textbf{ERCCFP(FN)(TP)IoB:} Ego-relative \commandcenter\ false positive/negative/true positive increment over \commandcenter\ baseline.
    \item \textbf{ERCCFP(FN)(TP)IoE:} Ego-relative \commandcenter\ false positive/negative/true positive increment over ego's local.
\end{itemize}

\subsubsection{Results}

In this work, we focus on presenting the ideals of using \mast\ for flexible security analysis. Thus, we leave a more rigorous investigation into the MC results for future works. Many more parameters are available for MC analysis including the target selection parameters, adversary settings, ego/agent sensor fusion pipelines, and \commandcenter\ integrity algorithms.

Results are distilled into Table~\ref{tab:monte-carlo}. Generally, attacks induce more false positives than false negatives. This is primarily due to the inherent vulnerability of multi-agent fusion to isolated false positives. If a single agent generates a longitudinally-consistent false positive, it can propagate into a track because it is unknown whether other agents could view that object to confirm/deny it; i.e.,~they could simply miss the object because of occlusion. It is difficult to distinguish between ``natural misses'' and ``adversarial misses''. On the other hand, the effect on false negatives is minimal. This supports the inherent resiliency of multi-agent fusion to isolated false negatives. Collaborative agents with views of the object can compensate for object negation in a single agent.

\newcommand{\panelonesubcaption}{Ego-relative command center false negative/false positive/true positive increment over baseline/ego-local outcome. Compares \commandcenter\ operating picture under attack scenario to \commandcenter\ under unattacked baseline (IoB) and to ego's local information (IoE).}
\newcommand{\paneltwosubcaption}{Compromised agent false negative/positive increment over baseline for each agent.}

\begin{table*}[t]
\centering

\label{tab:monte-carlo}
\begin{subtable}[t]{\linewidth}
\centering
\begin{tabular}{lrlllllll}
\toprule
Run ID & Coord? & \# Adv & ERCCFNIoB & ERCCFNIoE & ERCCFPIoB & ERCCFPIoE & ERCCTPIoB & ERCCTPIoE \\ \midrule
\midrule
C-1 & True & 1 & 0.00 +/- 0.00 & -7.60 +/- 2.67 & 3.21 +/- 1.47 & 3.04 +/- 2.15 & 0.00 +/- 0.00 & 7.60 +/- 2.67 \\ \midrule
C-2 & True & 2 & 0.08 +/- 0.28 & -7.51 +/- 2.69 & 5.65 +/- 2.73 & 5.48 +/- 3.13 & -0.08 +/- 0.28 & 7.51 +/- 2.69 \\ \midrule
C-3 & True & 3 & 0.10 +/- 0.29 & -7.50 +/- 2.68 & 5.74 +/- 2.63 & 5.56 +/- 3.03 & -0.10 +/- 0.29 & 7.50 +/- 2.68 \\ \midrule
UC-1 & False & 1 & 0.01 +/- 0.11 & -7.58 +/- 2.68 & 7.92 +/- 2.76 & 7.74 +/- 3.19 & -0.01 +/- 0.11 & 7.58 +/- 2.68 \\ \midrule
UC-2 & False & 2 & 0.27 +/- 0.47 & -7.32 +/- 2.78 & 15.67 +/- 5.29 & 15.49 +/- 5.49 & -0.27 +/- 0.47 & 7.32 +/- 2.78 \\ \midrule
UC-3 & False & 3 & 0.33 +/- 0.50 & -7.26 +/- 2.76 & 23.30 +/- 8.19 & 23.12 +/- 8.26 & -0.33 +/- 0.50 & 7.26 +/- 2.76 \\ \midrule
\bottomrule
\end{tabular}
\vspace{1pt}
\caption{\panelonesubcaption}
\end{subtable}
\begin{subtable}[t]{\linewidth}
\centering
\begin{tabular}{lrlllllll}
\toprule
Run ID & Coord? & \# Adv & CAFNIoB\_agent1 & CAFNIoB\_agent2 & CAFNIoB\_agent3 & CAFPIoB\_agent1 & CAFPIoB\_agent2 & CAFPIoB\_agent3 \\ \midrule
\midrule
C-1 & True & 1 & 0.00 +/- 0.00 & N/A & N/A & 3.21 +/- 1.47 & N/A & N/A \\ \midrule
C-2 & True & 2 & 0.08 +/- 0.28 & 0.08 +/- 0.28 & N/A & 5.65 +/- 2.73 & 5.65 +/- 2.73 & N/A \\ \midrule
C-3 & True & 3 & 0.10 +/- 0.29 & 0.10 +/- 0.29 & 0.10 +/- 0.29 & 5.74 +/- 2.63 & 5.74 +/- 2.63 & 5.74 +/- 2.63 \\ \midrule
UC-1 & False & 1 & 0.01 +/- 0.11 & N/A & N/A & 7.92 +/- 2.76 & N/A & N/A \\ \midrule
UC-2 & False & 2 & 0.27 +/- 0.47 & 0.27 +/- 0.47 & N/A & 15.67 +/- 5.29 & 15.67 +/- 5.29 & N/A \\ \midrule
UC-3 & False & 3 & 0.33 +/- 0.50 & 0.33 +/- 0.50 & 0.40 +/- 1.13 & 23.30 +/- 8.19 & 23.30 +/- 8.19 & 23.59 +/- 7.85 \\ \midrule
\bottomrule
\end{tabular}
\vspace{1pt}
\caption{\paneltwosubcaption}
\end{subtable}

\caption{Results over 6 runs of (un)coordinated attacks with different numbers of adversaries. Metrics described in Section~\ref{sec:6-evaluation-mc} and illuminate agent/\commandcenter\ performance changes over both unattacked baseline and local-only information. Effect on false positives is much larger than false negatives. Multi-agent fusion is inherently vulnerable to isolated false positives propagating into fake objects and inherently resilient to isolated instances of false negatives when other agents can provide support.}
\end{table*}
\section{\quad Limitations and Planned Improvements}

\myparagraph{Perception algorithms.} To emphasize the utility of the general framework, we bypassed perception algorithms online and provided detections to agents that were generated offline. Perception models are included as a part of our pipelining framework and should be included in future evaluations.

\myparagraph{Command center integrity.} We did not consider security-aware processing at the command center in the form of integrity algorithms. An investigation into the effectiveness of different schemes of centralized data integrity is a research endeavor in itself and is left to future works. 

\myparagraph{Mobile agents.} We restricted the number of mobile agents based on the publicly released CARLA multi-agent dataset. Given the released code for generating new datasets in CARLA for multi-agent research, we plan to generate richer datasets for future analysis with multiple mobile agents.

\myparagraph{Mission-critical planning and control.} The goal of the attacker was to maximally deteriorate the perspective at the command center. Commensurate with this objective, our analysis stopped at the level of the \commandcenter. Future works will fold planning and control algorithms in to this framework to assess the impact of attacks on safety and mission-critical algorithms.
\section{\quad\ Conclusion}

We presented a multi-agent security testbed, \mast, as a scalable framework built on ROS for security analysis of multi-agent, collaborative sensor fusion. \mast\ leverages recently-released open-source software for multi-agent algorithms~\cite{avstack} and dataset generation~\cite{hallyburton2023datasets}. \mast\ sets configuration at launch-time rather than build-time to enable diverse and rapid prototyping of AVs and manipulation of adversaries for security analysis with simple, high-level python code. We described through case studies and Monte Carlo results the effects of (un)coordinated adversaries on centralized fusion of multi-agent perception data to motivate the need for research into \msma\ integrity algorithms.


\bibliographystyle{IEEEtranS}
\bibliography{references}

\end{document}